\documentclass[lettersize,journal]{IEEEtran}
\usepackage{amsmath,amsfonts}
\usepackage{algorithmic}
\usepackage{array}
\usepackage[caption=false,font=normalsize,labelfont=sf,textfont=sf]{subfig}
\usepackage{textcomp}
\usepackage{stfloats}
\usepackage{url}
\usepackage{verbatim}
\usepackage{graphicx}
\def\BibTeX{{\rm B\kern-.05em{\sc i\kern-.025em b}\kern-.08em
    T\kern-.1667em\lower.7ex\hbox{E}\kern-.125emX}}
\usepackage{balance}

\usepackage{doi}
\usepackage{microtype}
\usepackage{multirow}
\usepackage[table,xcdraw]{xcolor}

\usepackage{placeins}
\usepackage{afterpage}
\usepackage[ruled,vlined]{algorithm2e}
\definecolor{mycolor}{RGB}{0, 0, 0}
\definecolor{mycolorR2}{RGB}{0, 0, 0}
\usepackage{hyperref}

\begin{document}

\title{Context-aware knowledge graph framework for traffic speed forecasting using graph neural network}
\author{Yatao Zhang, Yi Wang, Song Gao, Martin Raubal
\thanks{The research was conducted at the Future Resilient Systems at the Singapore-ETH Centre, which was established collaboratively between ETH Zurich and the National Research Foundation Singapore. This research is supported by the National Research Foundation Singapore (NRF) under its Campus for Research Excellence and Technological Enterprise (CREATE) programme. \textit{(Corresponding author: Yatao Zhang)}}
\thanks{Yatao Zhang and Martin Raubal are with Future Resilient Systems, Singapore-ETH Centre, ETH Zurich, Singapore 138602, Singapore; and Institute of Cartography and Geoinformation, ETH Zurich, Zurich 8093, Switzerland. (e-mail: yatzhang@ethz.ch; mraubal@ethz.ch)}
\thanks{Yi Wang is with Civil and Natural Resources Engineering Department, University of Canterbury, Christchurch 8041, New Zealand. (e-mail: y.wang@canterbury.ac.nz)}
\thanks{Song Gao is with Geospatial Data Science Lab, Department of Geography, University of Wisconsin-Madison, Madison, WI 53706, USA. (e-mail: song.gao@wisc.edu)}
}
\maketitle

\begin{abstract}
Human mobility is intricately influenced by urban contexts spatially and temporally, constituting essential domain knowledge in understanding traffic systems. While existing traffic forecasting models primarily rely on raw traffic data and advanced deep learning techniques, incorporating contextual information remains underexplored due to insufficient integration frameworks and the complexity of urban contexts.
This study proposes a novel context-aware knowledge graph (CKG) framework to enhance traffic speed forecasting by effectively modeling spatial and temporal contexts. Employing a relation-dependent integration strategy, the framework generates context-aware representations from the spatial and temporal units of CKG to capture spatio-temporal dependencies of urban contexts. A CKG-GNN model, combining the CKG, dual-view multi-head self-attention (MHSA), and graph neural network (GNN), is then designed to predict traffic speed utilizing these context-aware representations.
Our experiments demonstrate that CKG's configuration significantly influences embedding performance, with ComplEx and KG2E emerging as optimal for embedding spatial and temporal units, respectively. 
The CKG-GNN model establishes a benchmark for 10-120 min predictions, achieving average MAE, MAPE, and RMSE of 3.46±0.01, 14.76±0.09\%, and 5.08±0.01, respectively. Compared to the baseline DCRNN model, integrating the spatial unit improves the MAE by 0.04 and the temporal unit by 0.13, while integrating both units further reduces it by 0.18.
The dual-view MHSA analysis reveals the crucial role of relation-dependent features from the context-based view and the model's ability to prioritize recent time slots in prediction from the sequence-based view.
Overall, this study underscores the importance of merging context-aware knowledge graphs with graph neural networks to improve traffic forecasting.
\end{abstract}


\section{Introduction}
Urban transportation is significantly influenced by surrounding environments due to the supply-demand dynamics of human mobility in the urban space \cite{wegener2021land}. This environment-related influence manifests across both spatial and temporal contexts, which can benefit the predictive models for traffic speed \cite{zhang2023incorporating, shaygan2022traffic}. Actually, context awareness has emerged as a valuable tool in intelligent transportation systems (ITS) to adapt to various traffic scenarios by fully utilizing multi-source context datasets \cite{peng2019knowledge}. Despite its potential on ITS, developing context-aware models to boost traffic forecast and investigate contextual effects remains under exploration.

Short-term traffic state prediction involves forecasting upcoming traffic conditions on road segments for periods within hours \cite{yin2021deep}. 
Recent developments have predominantly focused on advanced deep learning methods, with graph neural networks standing out by leveraging spatio-temporal dependencies of traffic data \cite{tedjopurnomo2020survey}.
However, human mobility occurs in situation-dependent settings, significantly influenced by external factors such as land use and transportation networks \cite{buchin2012context}. These factors constitute the broader context that permeates urban environments, shaping and constraining individual movements across road segments \cite{zhang2022street}. For example, points of interest (POIs) and land use data illustrate the spatial arrangement of critical locations and facilities, enhancing our understanding of the environment surrounding transportation networks; traffic jams and weather data provide real-time insights into congestion and meteorological events, respectively, aiding in traffic prediction under various conditions.
Despite the critical role of contextual information, existing studies have largely centered on utilizing raw traffic data alongside deep learning techniques for forecasting tasks \cite{shaygan2022traffic}, neglecting the rich contextual information that could enhance predictions in complex urban environments.
Therefore, there is significant potential in developing context-based neural-symbolic methods to further improve traffic forecasting by integrating graph neural networks with knowledge representation techniques \cite{zhang2023incorporating,yu2023survey}.

Despite the potential benefits of using surrounding information in traffic forecasting, the context datasets carrying related knowledge are diverse and complex, deepening the difficulty in integrating them into the prediction task \cite{tedjopurnomo2020survey}. Generally, urban contexts influencing the traffic system encompass various spatial and temporal factors with a broad scope.
From a spatial perspective, the morphological layout of a city impacts how transportation systems are designed and operated \cite{wegener2021land, kang2012intra}, which in turn affects human mobility and traffic patterns. Spatial datasets, such as POIs, land uses, and transportation networks, are typical examples that describe the spatial dimension of urban contexts \cite{shaygan2022traffic, zhang2022street}. 
Temporally, the traffic patterns vary significantly throughout time periods, weather conditions, and traffic status \cite{yin2021deep}. For example, peak hours, typically during morning and evening commutes, require a transportation system capable of handling a high volume of passengers. Conversely, during off-peak hours, the system still needs to operate efficiently but with potentially less demand. 
Regardless of multiple sources, modalities, and dimensions, these spatio-temporal datasets are vital in shaping the nature of urban transportation, becoming a pivotal role in traffic speed prediction \cite{zhang2023incorporating}. 
Therefore, effectively organizing context datasets is the prerequisite to incorporating them into the prediction task.

In an effort to organize and facilitate the assimilation of comprehensive traffic-related information, knowledge graphs (KG) emerge as a valuable tool that has attracted substantial attention in the transportation field \cite{wang2021spatio, zhang2023mobility, ahmed2022knowledge}. The capacity to model relations and domain information makes knowledge graphs powerful in extracting high-level representations from different context datasets by utilizing embedding techniques \cite{dai2020survey}. 
However, applying knowledge graphs to traffic forecasting encounters two substantial challenges, i.e., (1) constructing context-aware knowledge graphs associating traffic-related contexts spatially and temporally, and (2) embedding the constructed knowledge graphs and generating context-aware representations spatially and temporally. These two challenges are further deepened considering the diversity and complexity of context datasets. 
In addition, deep learning architectures, such as graph neural networks (GNN), have become mainstream in traffic forecasting \cite{tedjopurnomo2020survey, yin2021deep, shaygan2022traffic}, but integrating context representations derived from knowledge graphs with these deep learning models to enhance prediction accuracy is still under exploration.
Meanwhile, integrating knowledge graphs into GNNs introduces additional parameters, potentially increasing the model's complexity, but the domain-specific insights provided by the knowledge graph can enhance the model's learning efficiency. This can be beneficial in scenarios with limited traffic data, where the structured information from the knowledge graph helps the model to make more informed predictions, leveraging contextual understanding to compensate for data scarcity.

To address these challenges, we propose a context-aware knowledge graph (CKG) framework to effectively model and embed the spatio-temporal relationships inherent in diverse urban contexts. This framework facilitates the integration of various context datasets into machine-readable formats. Then, we strategically combine the proposed CKG framework with GNN through dual-view multi-head self-attention (MHSA) techniques to forecast traffic speed. This integration ensures that CKG-based context representations can be effectively incorporated into traffic forecasting models, leveraging the strengths of both knowledge graphs and advanced neural network architectures to improve predictive performance.
Overall, the contributions of this study are three-fold:
\begin{itemize}
    \item Propose a context-aware knowledge graph (CKG) framework to model the spatio-temporal relations of context datasets using domain-specific information in the transportation field;
    \item Design a relation-dependent integration strategy to generate context-aware representations for traffic speed forecasting. Also, we investigate the performance of various KG embedding techniques and parameter configurations in modeling spatial and temporal contexts;
    \item Introduce a dual-view MHSA method to integrate the proposed CKG framework with GNN for traffic speed prediction, i.e., CKG-GNN, assisting in understanding how contexts affect traffic forecasting from the contextual and sequential views.
\end{itemize}

\section{Related work}
\subsection{Knowledge graphs in transportation}
Knowledge graphs refer to multi-relational graphs linking various entities through relations, which can structure complex datasets to represent and manage knowledge across domains effectively \cite{chen2020review, cao2024knowledge}.
These graph-based structures enable the modeling of domain-specific knowledge by organizing datasets into a network of interconnected entities and their relations, thus facilitating comprehensive semantic representations \cite{wang2017knowledge}. 
Hence, knowledge graphs are invaluable in applications ranging from ITS to smart city initiatives, where they contribute significantly to the modeling of urban dynamics and environments \cite{zhang2024knowledge}. For example, projects like the CARES program have demonstrated the potential of integrating dynamic geospatial knowledge graphs with semantic 3D city databases and agent-based systems for intelligent automation \cite{chadzynski2022semantic}.

The application of knowledge graphs in the transportation sector has particularly focused on two promising avenues. 
The first avenue is to establish urban knowledge graphs of movement datasets to capture the spatio-temporal relationships of human mobility directly \cite{zhuang2017understanding}, such as converting the task of human mobility prediction into a knowledge graph completion problem. In detail, Wang et al. \cite{wang2021spatio} introduced a spatio-temporal urban knowledge graph (STKG) to extract structured knowledge from massive trajectory data and then predicted human's future movement based on the extracted knowledge from spatio-temporal mobility patterns.
Second, we can employ knowledge graphs to build relations of context datasets and integrate their representations into traffic forecasting \cite{peng2019knowledge, zhang2023mobility}. A representative example is that Zhu et al. \cite{zhu2022kst} constructed a specially designed knowledge graph to encode external factors (e.g., POIs, weather conditions, and time) and then input them into a graph convolutional network to predict traffic speed. 
Basically, the incorporation of context-aware knowledge graphs into traffic forecasting remains preliminary, with the representation of complex spatio-temporal dependencies posing a significant challenge \cite{zhu2022kst, zhang2023mobility}.
Knowledge graph embedding (KGE) techniques provide a promising solution to the challenge by converting logic expressions of entities and relations within knowledge graphs to machine-readable representations \cite{dai2020survey}. 

Essentially, KGE techniques transform discrete graph structures into a continuous vector space, where entities and relations within a knowledge graph are encoded as vector representations \cite{dai2020survey, rossi2021knowledge}, thereby facilitating their integration into traffic forecasting tasks. This process involves defining a scoring function for each fact to ascertain its plausibility, i.e., a triple of (head entity, relation, tail entity), allowing for the quantitative assessment of the relations and entities within the knowledge graph \cite{cao2024knowledge}.
Wang et al. \cite{wang2017knowledge} categorized KGE approaches into two primary types, i.e., translational distance models (distance-based) and semantic matching models (similarity-based). 
Translational distance models, including TransE and TransR, evaluate the plausibility of facts by measuring the distance between entities after being translated by the relation vector \cite{chen2020review}. An adaptation, KG2E, takes uncertainty into account by modeling entities and relations as vectors from multivariate Gaussian distributions \cite{he2015learning}.
Semantic matching models, on the other hand, rely on similarity-based scoring functions. They determine the plausibility of facts by matching the latent semantics of entities and relations in their vector space embeddings. Examples of this approach include RESCAL, ComplEx, and NTN, which leverage the interplay of vector space dimensions to encode relational patterns \cite{wang2017knowledge}.
Additionally, several advanced variants of KGE techniques incorporate extra information into the embeddings, such as entity types, relation paths, textual descriptions, and logical rules \cite{chen2020review, wang2017knowledge}. These kinds of variants can enrich the representational capacity of knowledge graphs by considering their spatio-temporal dependencies. 
In this study, we aim to design a context-aware knowledge graph for traffic speed forecasting considering the spatio-temporal relations of available context datasets. Furthermore, we develop a relation-dependent integration strategy to capture the spatio-temporal path dependencies of urban contexts.

\subsection{Deep learning in traffic forecasting}
Traffic forecasting serves as a fundamental component in enhancing transportation resilience, which plays a crucial role in alleviating congestion and facilitating ITS's development \cite{kumar2021applications}. 
The scope of traffic forecasting encompasses various indicators, including traffic flow, speed, density, and demand, each offering unique but interconnected insights into the status of transportation networks \cite{yin2021deep, feng2023macro}. Despite different indicators, the goal of traffic prediction is to forecast traffic states based on historical traffic data and external factors.
Many methods have been developed to meet this forecasting objective, ranging from conventional statistical approaches to advanced machine learning and deep learning techniques \cite{shaygan2022traffic}. Notably, deep learning models have gained significant attention for their exceptional ability to handle high-dimensional datasets and capture intricate spatial-temporal dependencies inherent in traffic data \cite{tedjopurnomo2020survey}.

So far, different variants of deep learning models have emerged to forecast traffic speed, including convolutional neural networks (CNN), recurrent neural networks (RNN), and graph neural networks (GNN). 
CNN-based methods are pivotal for grid-based prediction tasks by leveraging their proficiency in processing spatial data and analyzing traffic patterns within grid-like structures \cite{ kumar2021applications}. However, applying CNN-based models to traffic forecasting is constrained due to the graph structure of transportation networks. Meanwhile, the inconvenience of modeling temporal dependencies in sequences also severely limits its performance in the prediction task \cite{shaygan2022traffic}. 
In contrast, RNN-based models, such as long short-term memory (LSTM) units and gated recurrent units (GRU), have been proposed to capture the temporal dependencies inherent in traffic data, a feat less attainable by conventional machine learning techniques \cite{kumar2021applications}. Despite their potential in modeling time-series data, RNN-based models fail to elucidate the spatial dependencies in transportation networks since traffic status in a given road is significantly affected by its neighboring roads \cite{shaygan2022traffic}.
The development of GNN-based models represents the forefront of efforts to address the need for capturing both spatial and temporal dependencies in traffic forecasting \cite{yin2021deep, feng2023macro}.
GNNs are neural architectures that process data represented as graphs by aggregating and updating node features through their connections, making them well-suited for tasks related to transportation networks \cite{tedjopurnomo2020survey}. Variants such as the diffusion convolution recurrent neural network (DCRNN) \cite{li2017diffusion}, spatial-temporal graph convolutional network (STGCN) \cite{yu2017spatio}, and temporal graph convolutional network (TGCN) \cite{zhao2019t} are some typical examples of utilizing GNNs effectively for traffic forecasting. These models underpin the current study's approach to predicting traffic status, showcasing the advanced capabilities of GNNs in this domain.

Furthermore, urban contexts are also important for traffic forecasting by influencing and shaping human mobility patterns \cite{buchin2012context, zhang2022street}. Tedjopurnomo et al. \cite{tedjopurnomo2020survey} have underscored the critical role of contextual information in enhancing traffic forecasting, highlighting the benefits of integrating context-aware insights into predictive models. Nowadays, context-aware modeling is increasingly gaining prominence in the field of traffic forecasting.
For example, Zhu et al. \cite{zhu2021ast} proposed an attribute-augmented spatio-temporal graph convolutional network (AST-GCN) to model external factors as dynamic and static attributes for traffic prediction. 
Zhang et al. \cite{zhang2023incorporating} specified multimodal contexts from spatial and temporal views, and then proposed a multimodal context-based graph convolutional neural network (MCGCN) model to fuse context datasets into traffic prediction. The MCGCN model employs two distinct approaches for managing spatial and temporal contexts: hierarchical spatial embedding, which generates spatial representations by organizing spatial contexts across various dimensions, and multivariate temporal modeling, which learns temporal representations by capturing the dependencies within multivariate temporal contexts.
These examples have verified the superiority of using contextual information in traffic forecasting compared to existing GNN-based models.
Generally, the integration of context-aware strategies into deep learning models represents a pivotal development in traffic forecasting, which lays a robust foundation for this study's focus on speed prediction by leveraging knowledge graphs.
Building upon existing context-aware strategies, our study aims to propose a CKG framework that utilizes knowledge graphs for modeling domain-specific information in the transportation field, and then integrates this framework into GNNs to improve traffic speed forecasting.

\section{Methodology}
\subsection{Problem statement}
\begin{figure}[htbp!]
  \centering
  \includegraphics[width=0.95\linewidth]{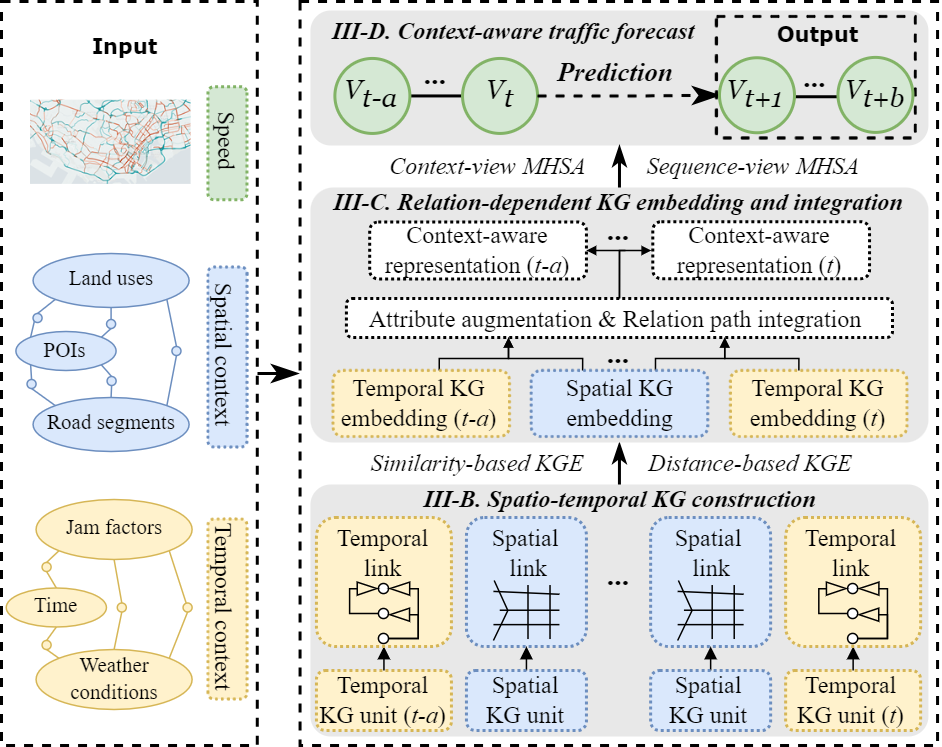}
  \caption{An overview of the CKG-GNN model for traffic speed forecasting. It comprises three parts: spatio-temporal KG construction, relation-dependent KG embedding and integration, and context-aware traffic forecast. The model processes inputs of traffic speed, spatial context, and temporal context data from the past $a$ time steps, and outputs traffic speeds for the forthcoming $b$ time steps.}
  \label{fig:framework}
\end{figure}

Assuming $V$, $S$, and $T$ represent raw traffic speed at road segments, spatial context datasets, and temporal context datasets, respectively, the traffic forecast problem is defined as predicting traffic speed in the next $b$ time slots utilizing $V$, $S$, and $T$ in the past $a$ time slots. 
To solve this problem, this study proposes a CKG framework for traffic forecasting using GNN, i.e., a CKG-GNN, to construct context-aware knowledge graphs and then incorporate the context-aware representations into traffic speed prediction. As shown in Figure \ref{fig:framework}, the proposed CKG-GNN model consists of three parts: (1) spatio-temporal knowledge graph construction, (2) relation-dependent knowledge graph embedding and integration, and (3) context-aware traffic forecast using dual-view MHSA.
The code and test datasets of CKG-GNN are available at https://github.com/mie-lab/CKG-traffic-forecasting.

\subsection{Spatio-temporal knowledge graph construction}
A knowledge graph is a fact-composed graph linking entities through relations. A fact, i.e., an edge in the graph, is expressed as a triple $(h, r, t)$, where $h$, $r$, and $t$ are the head entity, relation, and tail entity, respectively. All observed facts constitute the context-aware knowledge graph, denoted as $\mathbb{K}$. We construct the CKG $\mathbb{K}$ encompassing two units, i.e., a spatial unit $\mathbb{K}_S$ and a temporal unit $\mathbb{K}_T$, organizing spatial and temporal contexts influencing traffic forecasting, respectively. By using the CKG, we can represent different contexts with the same format to store domain knowledge for the subsequent traffic forecasting task.

\subsubsection{Spatial unit construction}\label{sec:spat_unit}
The spatial unit $\mathbb{K}_S$ organizes spatial context datasets by leveraging the interconnected spatial relations among context entities that influence traffic forecasting.
In the transportation domain, spatial contexts affecting traffic status and human mobility are diverse. We select three representative spatial contexts according to existing studies, including POIs, road segments, and land uses \cite{buchin2012context, zhu2022kst, zhang2023incorporating}. These three factors represent spatial contexts from varied spatial dimensions, i.e., spatially-discrete points, spatially-linked lines, and spatially-continuous planes, respectively \cite{zhang2023incorporating}.
Assuming the set of POIs as $P^s = \{p_1, ..., p_i, ..., p_{|P^s|}\}$, the set of road segments as $R^s=\{s_1, ..., s_i, ..., s_{|R^s|}\}$, and the set of land uses as $L^s=\{l_1, ..., l_i, ..., l_{|L^s|}\}$, we establish CKG's spatial unit $\mathbb{K}_S$ using the following three kinds of facts:
\begin{itemize}
    \item $\mathbb{K}_S(road)$: $(road, adjacentToRoad, road)$. Given $s_i, s_j \allowbreak \in R^s$, this fact represents whether road $s_i$ is adjacent to road $s_j$. Also, we attached an attribute fact to $\mathbb{K}_S(road)$ for enriching the semantic information of road entities, namely $(road, hasFFSpeed, speed)$, which provides the free flow speed of the given road. This expression allows us to model the spatial relationship between roads in the CKG, taking roads' baseline conditions into account.
    \item $\mathbb{K}_S(poi)$: $(poi, locatedInBuffer[Dist], road)$ and $(poi \allowbreak , hasType, poiType)$. These two facts associate road $s_i$ with different types of POIs in $P^s$ considering their distance to $s_i$. $poiType$ refers to the POI category. $[Dist]$ is the distance to create a buffer for road $s_i$ to identify whether a given POI $p_i$ is located in the created buffer, as the distances of POIs to a road affect their context effect on human mobility \cite{zhang2022street}. Here, we construct two buffer types to investigate this distance effect: $[Dist]$ from 10 to 100 meters with an interval of 10 meters and $[Dist]$ from 100 to 500 meters with an interval of 100 meters. Also, we have an attribute fact for $\mathbb{K}_S(poi)$, namely $[road, hasPoi[Type]InBuffer[Dist], poiCount]$, meaning the number of POIs with $[Type]$ in $s_i$'s buffer with $[Dist]$.
    \item $\mathbb{K}_S(land)$: $(land, intersectWithBuffer[Dist], road)$ and $(land, hasType, landType)$. The $\mathbb{K}_S(land)$ fact is similar to $\mathbb{K}_S(poi)$ except for changing the spatial relation from $locatedInBuffer[Dist]$ to $intersect \allowbreak WithBuffer[Dist]$. Instead, the attribute fact is $[road,  \allowbreak hasLand[Type]InBuffer[Dist], AreaRatio]$, meaning the intersected area ratio of land $l_i$ with $[Type]$ to $s_i$'s buffer with $[Dist]$. $landType$ is the land category. 
\end{itemize}

To further utilize the spatial topology information of the transportation network, we introduce a spatial link fact that captures the adjacency order between any two road segments $s_i$ and $s_j$ as follows:
\begin{itemize}
    \item $\mathbb{K}_S(link)$: $(road, spatiallyLink[Order], road)$. The fact establishes a connection between $s_i$ and $s_j$, where $[Order]$ denotes the number of hops between the two roads, with the range capped at 6. For example, an $[Order]$ of 1 implies direct adjacency between $s_i$ and $s_j$; an $[Order]$ of 2 indicates that the roads are connected via an intermediate road. This fact is designed as it acknowledges that the traffic condition on a given road is influenced not just by its immediate neighbors but also by roads within a broader adjacency context.
\end{itemize}

\subsubsection{Temporal unit construction}\label{sec:temp_unit}
The temporal unit $\mathbb{K}_T$ exemplifies the relations of temporal context datasets in the traffic system, accounting for their temporal changes. 
In the transportation domain, various time-sensitive factors play a crucial role in traffic forecasting, such as time indicators that mark variations in traffic speed at different times of the day or on different days of the week \cite{kumar2021applications}. 
To augment $\mathbb{K}_T$, we also include traffic jam factors and weather conditions due to their significant influence on traffic dynamics: traffic jam factors are pivotal for tracking the fluctuating nature of vehicle speed and the immediate effect of traffic congestion, and weather information indicates driving behaviors and conditions \cite{tedjopurnomo2020survey, yin2021deep}.
For the set of roads $R^s$, we define $T^t = \{hour_t, day_t\}$ to represent the hour of the day and the day of the week at time $t$. $J^t = \{j^t_1, ..., j^t_i, ..., j^t_{|R^s|}\}$ refers to the jam factors for each road segment in $R^s$ at time $t$, and $W^t = \{w^t_1, ..., w^t_i, ..., w^t_{|R^s|}\}$ denotes the corresponding weather conditions for these road segments at time $t$. These factors contribute to the construction of the temporal unit $\mathbb{K}_T$ through the following facts:
\begin{itemize}
    \item $\mathbb{K}_T(time)$: $(road, hasHour, hour)$ and $(road, hasDay, \allowbreak day)$. The facts attach time indicators $t$, i.e., the hour of the day and the day of the week, to the given road $s_i$. Their attributes reflect human mobility's difference in various hours ranging [1, 24] and days ranging [1, 7]. A cosine function method is applied to highlight the periodic characteristics of time indicators, with $cosine(2\pi \times hour/24)$ for hours and $cosine(2\pi \times day/7)$ for days.  
    \item $\mathbb{K}_T(jam)$: $(road, hasJam[PastMins], jam)$. Given a road $s_i$, this fact captures its average traffic congestion status over a past duration at time $t$. $[PastMins]$ refers to the past minutes, ranging from 10 to 60 minutes with an interval of 10 minutes. The corresponding attribute $jam$ averages jam factors within $[PastMins]$ minutes. Similar to the spatial buffer in $\mathbb{K}_S(poi)$ and $\mathbb{K}_S(land)$, $[PastMins]$ represents the time buffer for time $t$. This relation helps understand the recent congestion trends on the road to boost the prediction of future traffic conditions.
    \item $\mathbb{K}_T(weather)$: $(road, hasTprt[PastMins], tprt)$, $(road, hasRain[PastMins], rain)$, and $(road, has \allowbreak Wind \allowbreak [PastMins], wind)$. $\mathbb{K}_T(weather)$ comprises three types of relations, each offering an average of the recent weather conditions within $[PastMins]$ minutes at time $t$, including $tprt$ for air temperature, $rain$ for rainfall, and $wind$ for wind speed. Their attribute values, derived from the nearest weather station data to the given road $s_i$, are crucial for assessing the impact of weather on traffic dynamics.
\end{itemize}

In addition, understanding the cyclic nature of traffic dynamics, manifesting on an hourly, daily, or weekly basis, can enhance traffic speed forecasting \cite{tedjopurnomo2020survey}. This cyclical nature reflects the routine behavior of human mobility, where traffic patterns tend to repeat at similar times on a daily basis (e.g., peak hours in the morning and evening) and weekly (increased traffic on weekends or weekdays). The hourly link is also included due to its direct influence on traffic conditions in the subsequent hour. To include this domain knowledge in $\mathbb{K}_T$, we define a temporal link fact that connects road entities with relevant temporal contexts:
\begin{itemize}
    \item $\mathbb{K}_T(link)$: $(road, temporallyLink[Temp][Link], [Tem \allowbreak p])$. Given a road $s_i$, $temporallyLink[Temp][Link]$ is designed to associate it with a specific temporal context $[Temp]$ through a link type $[Link]$. Here, $[Temp]$ could be time indicators derived from $\mathbb{K}_T(time)$, jam factors from $\mathbb{K}_T(jam)$, or weather conditions from $\mathbb{K}_T(weather)$. The $[Link]$ delineates the temporal link on an hourly, daily, or weekly basis, allowing for a thorough representation of time-related influences on traffic forecasting.
\end{itemize}

Meanwhile, the structure of $\mathbb{K}_T$ remains constant over time, while its attributes evolve due to the dynamics of temporal contexts. Moreover, real-time updates, such as sudden traffic jams or extreme weather events, solely influence the attributes without altering the structure of $\mathbb{K}_T$.
This setting avoids the need for re-embedding entities and relations within $\mathbb{K}_T$ at each time step in the prediction task or when including longer time periods, thereby improving efficiency while still capturing the essential temporal dynamics.

\subsection{Relation-dependent knowledge graph embedding and integration}
Given $\mathbb{K}$, the crucial task is transforming it into a format that deep learning models can process, enabling the incorporation of this knowledge into traffic forecasting. 
Knowledge graph embedding fundamentally involves mapping entities ($h$ and $t$) and relations ($r$) to a continuous vector space, i.e., $\textbf{h}, \textbf{t}, \textbf{r}$. This mapping is achieved by establishing a scoring function $f_r(h, t)$ to measure the plausibility of each fact $(h, r, t)$ \cite{wang2017knowledge}.
The goal is to generate the embeddings of $\textbf{h}$, $\textbf{r}$, and $\textbf{t}$ when the scoring function maximizes the total plausibility of the observed facts in $\mathbb{K}$, framing it as an optimization problem. Essentially, the model is designed to assign higher scores to facts that are present in $\mathbb{K}$, distinguishing them from non-observed facts.

\subsubsection{Knowledge graph embedding}\label{sec:kge}
KGE refers to embedding entities and relations of a knowledge graph into vector representations while preserving the inherent structure of the graph, which can be divided into two categories based on their scoring functions: distance-based and similarity-based models \cite{wang2017knowledge}. Distance-based models measure the plausibility of facts within $\mathbb{K}$ by calculating the distance between entities following a relation-specific translation. On the other hand, similarity-based models measure the plausibility by matching the latent semantics of entities and relations within their vector spaces. In our study, we delve into both model types, examining their efficacy in CKG embedding and choosing the most suitable methods for embedding $\mathbb{K}_S$ and $\mathbb{K}_T$.
To facilitate the understanding and comparison of distance-based and similarity-based models, Table \ref{tab:kge_tech} in the Appendix summarizes the scoring functions and key features of each knowledge graph embedding technique.

Regarding distance-based models, TransE serves as a foundational method by representing entities and relations as vectors within the same vector space $\mathbb{R}^e$ \cite{chen2020review}. For a given fact $(h, r, t)$, TransE's scoring function is defined as the negative distance between $\textbf{h} + \textbf{r}$ and $\textbf{t}$, i.e., 
\begin{equation}
    f_r(h, t) = -\left \| \textbf{h} + \textbf{r} - \textbf{t} \right \|^2_2
\end{equation}
where $\textbf{h}, \textbf{r}, \textbf{t} \in \mathbb{R}^e$. The score is expected to be higher if $(h, r, t) $ holds. However, TransE's simplistic approach can struggle with complex relation types like 1-to-N, N-to-1, and N-to-N due to its uniform treatment of entity and relation embeddings.

TransR addresses this limitation by introducing relation-specific spaces $\mathbb{R}^r$ \cite{chen2020review}. Given a fact $(h, r, t)$, TransR projects $\textbf{h}$ and $\textbf{t}$ to the relation-specific space through a projection matrix $\textbf{M}_r \in \mathbb{R}^{e \times r}$, i.e., $\hat{\textbf{h}}$ and $\hat{\textbf{t}}$. This projection enables TransR to effectively model diverse relationship types by computing the scoring function as follows:
\begin{equation}
    \hat{\textbf{h}} = \textbf{M}_r \textbf{h}, \,\,\,
    \hat{\textbf{t}} = \textbf{M}_r \textbf{t}, \,\,\,
    f_r(h, t) = -\left \| \hat{\textbf{h}} + \textbf{r} - \hat{\textbf{t}} \right \|^2_2
\end{equation}

KG2E advances the translational concept further by modeling entities and relations as random variables with multi-variate Gaussian distributions, rather than deterministic points in TransE and TransR \cite{he2015learning, chen2020review}. This variant enables KG2E to capture the inherent uncertainties in $\mathbb{K}$, with their representations shown in Equation~\ref{eq:kg2e}.
\begin{equation}
    \textbf{h} \sim \mathcal{N} (\boldsymbol{\mu}_h, \boldsymbol{\Sigma}_h), \,\,\,
    \textbf{r} \sim \mathcal{N} (\boldsymbol{\mu}_r, \boldsymbol{\Sigma}_r), \,\,\,
    \textbf{t} \sim \mathcal{N} (\boldsymbol{\mu}_t, \boldsymbol{\Sigma}_t)
    \label{eq:kg2e}
\end{equation}
Here, $\boldsymbol{\mu}_h, \boldsymbol{\mu}_r, \boldsymbol{\mu}_t \in \mathbb{R}^e$ are mean vectors, and $\boldsymbol{\Sigma}_h, \boldsymbol{\Sigma}_r, \boldsymbol{\Sigma}_t \in \mathbb{R}^{e\times e}$ are their corresponding covariance matrices.
Afterward, KG2E calculates the scoring function by measuring the probabilistic distance between the transformed $\textbf{t} - \textbf{h}$ and $\textbf{r}$, i.e., $ (\boldsymbol{\mu}_t - \boldsymbol{\mu}_h, \boldsymbol{\Sigma}_t + \boldsymbol{\Sigma}_h)$ and $(\boldsymbol{\mu}_r, \boldsymbol{\Sigma}_r)$. This distance is computed using the Kullback-Leibler divergence in Table \ref{tab:kge_tech} for the scoring function $f_r(h, t)$ to determine the plausibility of a fact.

In similarity-based models, RESCAL associates each entity with a vector and each relation with a matrix to model the pairwise interactions between latent factors using the bilinear function \cite{wang2017knowledge}. The scoring function of RESCAL is defined through the following form, i.e.,
\begin{equation}
    f_r(h, t) = \sum_{i=0}^{e-1}\sum_{j=0}^{e-1}[\textbf{M}_r]_{ij} \cdot [\textbf{h}]_i \cdot [\textbf{t}]_j
\end{equation}
where $\textbf{h}, \textbf{t} \in \mathbb{R}^{e}$ and $\textbf{M}_r \in \mathbb{R}^{e \times e}$. 
However, RESCAL's reliance on a matrix for each relation can lead to a surge in parameters, thereby elevating computational costs and storage demands.

ComplEx enhances the modeling capability by incorporating complex-valued embeddings, enabling the effective representation of asymmetric relationships \cite{cao2024knowledge}. Given a fact $(h, r, t)$, the obtained $\textbf{h}$, $\textbf{r}$, and $\textbf{t}$ are in a complex space, denoted as $\mathbb{C}^e$, and the scoring function is then defined as follows:
\begin{equation}
    f_r(h, t) = Real\left ( \sum_{i=0}^{e-1} [\textbf{r}]_i \cdot [\textbf{h}]_i \cdot [\bar{\textbf{t}}]_i \right )
\end{equation}
where $Real(\cdot)$ refers to taking the real part of the complex and $\bar{\textbf{t}}$ is the conjugate of $\textbf{t}$. This scoring function can capture the interaction between entities and relations, making ComplEx adept at handling asymmetric relations.

NTN (neural tensor network) introduces a tensor-based network architecture for embedding \cite{chen2020review}. The entities $\textbf{h}$ and $\textbf{t}$ are initially transformed into vector embeddings, i.e., $\textbf{h}, \textbf{t} \in \mathbb{R}^{e}$. Then, they are combined through a relation-specific tensor $\dot{\textbf{M}}_r \in \mathbb{R}^{e \times e \times r}$ and mapped to a non-linear hidden layer. Finally, a relation-specific output layer computes the score using Equation~\ref{eq:ntn}.
\begin{equation}
    f_r(h, t) = \textbf{\textbf{r}}^T \text{tanh}\left ( \textbf{h}^T \dot{\textbf{M}}_r \textbf{t} + \textbf{M}_r^1 \textbf{h} + \textbf{M}_r^2 \textbf{t} + \textbf{b}_r\right )
    \label{eq:ntn}
\end{equation}
Here, $\textbf{M}_r^1, \textbf{M}_r^2 \in \mathbb{R}^{e\times r}$ are relation-specific matrices and $\textbf{b}_r \in \mathbb{R}^{r}$ refers to bias vectors. Despite one of the most expressive models, NTN's demand for a substantial number of parameters during training poses challenges in terms of efficiency and computational resource requirements \cite{wang2017knowledge}.

To compare the difference of embedding $\mathbb{K}$ using different methods, we evaluate their performance on link prediction tasks using the PyKEEN library, where the goal is to predict missing relationships in a fact $(h, r, t)$ \cite{ali2021pykeen, rossi2021knowledge}. 
The evaluation is distinguished by the side of the prediction, i.e., (1) right-side: predicting $t$ using $h$ and $r$, (2) left-side: predicting $h$ using $r$ and $t$, and (3) both-side: combining both right-side and left-side evaluation. 
In this task, the commonly used evaluation metrics are based on the ranks of correct predictions, including mean rank (MR), mean reciprocal rank (MRR), and Hits@K (H@K) \cite{wang2017knowledge, rossi2021knowledge, cao2024knowledge}.
MR averages the ranks of correct entities within the prediction list; MRR calculates the average reciprocal rank of the first correct prediction; and H@K assesses the frequency of correct predictions within the top K results, such as H@5.
Moreover, to enhance evaluation accuracy by compensating for dataset-specific biases and imbalances, we have implemented the adjusted Hits@K (AH@K) metric, which considers the expectation of correct predictions appearing within the top K results \cite{hoyt2022unified}.
A lower MR indicates better performance, whereas higher values of MRR, H@K, and AH@K signify improved model performance.
When ranking the prediction list for computing evaluation metrics, we employ the realistic rank representing the expected value over all permutations respecting the sort order \cite{ali2021pykeen}.

To elucidate the process of CKG construction and embedding evaluation, we detail the process in Algorithm \ref{al:ckg_construct} using inputs of spatial context $S$ and temporal context $T$. The initial steps involve establishing the spatial unit $\mathbb{K}_S$ based on spatial facts in Section \ref{sec:spat_unit}, and the temporal unit $\mathbb{K}_T$ based on temporal facts in Section \ref{sec:temp_unit}. The framework then progresses to evaluate various KGE techniques to determine the most suitable for both $\mathbb{K}_S$ and $\mathbb{K}_T$ using evaluation metrics. The process concludes with the embedding results of $S$ and $T$.

\begin{algorithm}[tb]
\DontPrintSemicolon
\caption{CKG construction}
\label{al:ckg_construct}
\KwIn{spatialContext $S$, temporalContext $T$}
\KwOut{embedResults of $S$ and $T$}
INIT spatialUnit $\mathbb{K}_S$, temporalUnit $\mathbb{K}_T$\;
\For{\textnormal{spatialEntities in $S$}}{
    ADD spatialFacts $\mathbb{K}_S(road)$, $\mathbb{K}_S(poi)$, $\mathbb{K}_S(land)$ to $\mathbb{K}_S$\;
    ADD spatialLinkFact $\mathbb{K}_S(link)$ to $\mathbb{K}_S$\;
}
\For{\textnormal{temporalEntities in $T$}}{
    ADD temporalFacts $\mathbb{K}_T(time)$, $\mathbb{K}_T(jam)$, $\mathbb{K}_T(weather)$ to $\mathbb{K}_T$\;
    ADD temporalLinkFact $\mathbb{K}_T(link)$ to $\mathbb{K}_T$\;
}
\For{\textnormal{$\mathbb{K}$ in \{$\mathbb{K}_S$, $\mathbb{K}_T$\}}}{
    INIT bestMetrics, bestKGEmbed\;
    \For{\textnormal{KGEmbed in \{TransE, TransR, KG2E, RESCAL, ComplEx, NTN\}}}{
        EMBED $\mathbb{K}$ using KGEmbed\;
        COMPUTE \{MR, MRR, H@K, AH@K\}\;
        \If{\textnormal{\{MR, MRR, H@K, AH@K\} OUTPERFORM bestMetrics}}{
            bestMetrics = \{MR, MRR, H@K, AH@K\}\;
            bestKGEmbed = KGEmbed\;
        }
    }
    embedResults = EMBED $\mathbb{K}$ using bestKGEmbed\;
    \KwRet{\textnormal{embedResults}}\;
}
\end{algorithm}

\subsubsection{Relation-dependent integration for context representations}
After generating embeddings for each entity and relation in $\mathbb{K}_S$ and $\mathbb{K}_T$ using KGE techniques, we need to strategically deploy these embeddings to foster context-aware embeddings for the traffic forecasting task. 
Basically, each road in the transportation network represents a discrete unit encapsulating traffic speed data; thus, the simplest way is to incorporate road embeddings derived from $\mathbb{K}$ with speed data directly for the prediction.
Nevertheless, this method falls short of leveraging the full potential of $\mathbb{K}_S$ and $\mathbb{K}_T$, neglecting the graph structure and rich contextual information embedded within the spatial and temporal surroundings. The challenge lies in harnessing the comprehensive context embodied in CKG, which includes the complex spatial and temporal dependencies among various entities and their associated relations.

To solve this challenge, we propose a relation-dependent integration strategy that includes multi-hop relation path integration and attribute augmentation. This strategy is designed to capitalize on the position of road entities in CKG, thereby leveraging the inherent spatial-temporal relationship and context-specific information encoded within its graph structure.
A relation path is defined as a succession of linked relations forming a connective sequence between two entities, formally denoted as $r_1 \rightarrow \ldots \rightarrow r_l$ \cite{wang2017knowledge}. This multi-hop format enables us to trace and encode the relational trajectories within $\mathbb{K}_S$ and $\mathbb{K}_T$, capturing the full spectrum of associations surrounding each road segment. 
Given a relation path $p=r_1 \rightarrow \ldots \rightarrow r_l$ that adjoins a road segment to an entity $e$, and their respective embeddings notated as $\textbf{r}_1, \ldots, \textbf{r}_l$ and $\textbf{e}$, we define the relation-dependent embedding $\textbf{e}_{path}$ for $e$ in the following mathematical form:
\begin{equation}
    \textbf{e}_{path} = f_{[\text{dist}]}(\textbf{e} + \sum_{m=1}^{l}\textbf{r}_m ) + f_{[\text{sim}]}(\textbf{e} \cdot \prod_{m=1}^{l}\textbf{r}_m )
    \label{eq:rel_embed}
\end{equation}
Here, $f_{[\text{dist}]}$ and $f_{[\text{sim}]}$ represent indicator functions activated based on the use of distance-based or similarity-based KGE methods, respectively. $f_{[\text{dist}]}(\cdot)=1$ if a distance-based KGE method is utilized, and $f_{[\text{sim}]}(\cdot)=1$ if a similarity-based KGE method is applied. Otherwise, the indicator function defaults to zero. This bifurcation is due to the differing mechanisms by which distance-based and similarity-based models define their respective scoring functions in Section~\ref{sec:kge}.

The other component of the proposed relation-dependent integration strategy is attribute augmentation to enhance the semantic richness of each fact. This augmentation employs attribute facts associated with each fact in $\mathbb{K}_S$ and $\mathbb{K}_T$, such as $hasPoi[Type]InBuffer[Dist]$ for $\mathbb{K}_S(poi)$ and $hasLand[Type]InBuffer[Dist]$ for $\mathbb{K}_S(land)$, infusing additional attribute information into the embeddings.
For simplicity, we denote the attribute value for relations as $x$ and for entities as $y$. Consequently, the attribute-augmented embeddings for relations are $x_1\textbf{r}_1, ..., x_l\textbf{r}_l$, and for entities, it is $y\textbf{e}$. To ensure consistency and comparability across attributes, we apply max-min normalization to these attribute values separately, resulting in normalized values $x'$ and $y'$. Integrating these normalized attributes into the relation-dependent embedding yields the enriched context-aware embedding for $e$:
\begin{equation}
    \textbf{e}'_{path} = f_{[\text{dist}]}(y'\textbf{e} + \sum_{m=1}^{l}{x_m}'\textbf{r}_m) + f_{[\text{sim}]}(y'\textbf{e} \cdot \prod_{m=1}^{l}{x_m}'\textbf{r}_m)
    \label{eq:rel_embed_attr}
\end{equation}

\begin{figure}[htbp!]
  \centering
  \includegraphics[width=0.95\linewidth]{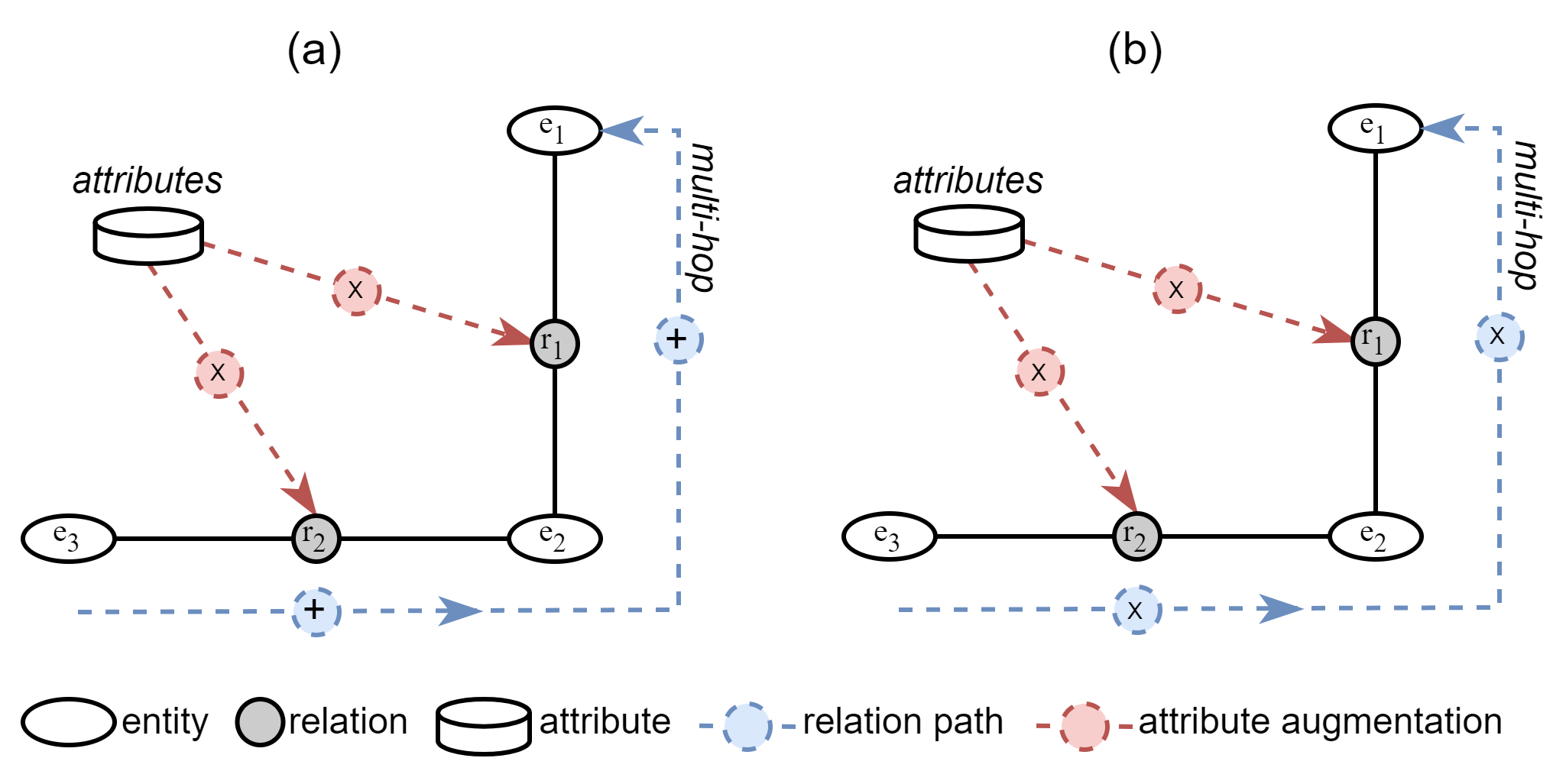}
  \caption{Diagrams illustrating multi-hop relation path integration and attribute augmentation for (a) distance-based models and (b) similarity-based models. The primary distinction between them lies in the method of relation path integration, with addition ($+$) employed in (a) and multiplication ($\times$) utilized in (b).}
  \label{fig:diag_multihop}
\end{figure}

Figure \ref{fig:diag_multihop} outlines our relation-dependent integration strategy by illustrating multi-hop relation path integration and attribute augmentation for both distance-based and similarity-based models.
In Figure \ref{fig:diag_multihop}(a), the objective is to generate the relation-dependent embedding for entity $\textbf{e}_1$ using a distance-based approach. To accomplish this, we identify a two-hop relation path $p_1$ associated with $\textbf{e}_1$, i.e., $p_1 = r_2 \rightarrow r_1$. The multi-hop relation path integration is then performed along $p_1$ using Equation \ref{eq:rel_embed}, where $f{[\text{dist}]}(\cdot)=1$ and $f{[\text{sim}]}(\cdot)=0$ signifies the application of distance-based models.
Meanwhile, attributes associated with each relation in the path, specifically $r_1$ and $r_2$, are incorporated into the final relation-dependent embedding using attribute augmentation, as per Equation \ref{eq:rel_embed_attr}.
Figure \ref{fig:diag_multihop}(b) outlines a similar process for similarity-based models, with the only difference being the function where $f_{[\text{dist}]}(\cdot)=0$ and $f_{[\text{sim}]}(\cdot)=1$, reflecting the use of similarity-based models.

It should be noted when embedding $\mathbb{K}_S$, we refine the representation of spatial contexts by directly linking road segments to $poiType$ and $landType$, bypassing the intermediate entities. This simplification, transforming $road \rightarrow poi \rightarrow poiType$ and $road \rightarrow land \rightarrow landType$ into direct $road \rightarrow poiType$ and $road \rightarrow landType$ associations, streamlines the graph structure. It can enhance the semantic integration of POIs and land uses, enabling more effective utilization of context data for the traffic prediction task.

\subsection{Context-aware traffic forecast using multi-head self-attention}
To integrate the context-aware embeddings derived from the CKG into the traffic forecasting model, i.e., CKG-GNN, a preliminary step involves combining these embeddings into a comprehensive representation. In this process, we concatenate the context-aware embeddings in Equation~\ref{eq:rel_embed_attr} from $\mathbb{K}_S$ and $\mathbb{K}_T$ according to the following formulation:
\begin{gather}    
    \textbf{e}'_S = \text{Concat}\left (\textbf{e}_{(road)_S}, \textbf{e}'_{\mathbb{K}_S(road; poi; land; link)} \right ) \\
    \textbf{e}'_T = \text{Concat}\left (\textbf{e}_{(road)_T}, \textbf{e}'_{\mathbb{K}_T(time; jam; weather; link)} \right )  \\
    \textbf{e}'_{fuse} = \text{Concat}\left (\textbf{e}'_S, \textbf{e}'_T \right )
\end{gather}
Here, $\textbf{e}_{(road)_S}$ and $\textbf{e}_{(road)_T}$ represent the road-only spatial and temporal features without relation-dependent integration, respectively.
$\textbf{e}'_{\mathbb{K}_S(road; poi; land; link)}$ refers to concatenating the spatial relation-dependent features of roads, POIs, lands, and spatial links. Similarly, $\textbf{e}'_{\mathbb{K}_T(time; jam; weather; link)}$ is to concatenate the temporal relation-dependent features of times, jam factors, weather conditions, and temporal links. 
The combined embedding $\textbf{e}'_{fuse}$ serves as the context-aware representation for the subsequent fusion process to capture the spatio-temporal dependencies.

Afterward, we utilize the multi-head self-attention (MHSA) mechanism to fuse embeddings from various contexts. The MHSA operates by executing parallel attention operations, emphasizing the focus on different representation segments simultaneously. It computes outputs by correlating queries with a set of key-value pairs, where queries $Q$, keys $K$, and values $V$ are vectors of equal dimensionality from the same source, namely $\textbf{e}'_{fuse}$. The core of the MHSA is represented by:
\begin{equation}
    \text{MultiHead}(Q, K, V) = \text{Concat}[head_1, \ldots, head_H]\boldsymbol{W}_0
    \label{eq:mhsa}
\end{equation}
\begin{equation}
    \text{where } head_i = \text{Attention}(Q\boldsymbol{W}_i^Q, K\boldsymbol{W}_i^K, V\boldsymbol{W}_i^V)
\end{equation}
In this study, we propose a novel method to apply MHSA from two views, i.e., context-based and sequence-based views, enabling a dual perspective analysis. Initially, the dual-view MHSA discerns the influence of various context features in $\mathbb{K}$, prioritizing those critical for forecasting from the context-based feature view. Subsequently, the sequence-based view employs the output from the context-based view to highlight the historical sequences vital for temporal prediction, emphasizing time slots that are most predictive for future traffic conditions. In addition, forward masking is employed in the sequence-based view to ensure temporal coherence by restricting attention to past and present data. This operation prevents the model from accessing future information, which is essential to avoid lookahead bias. 
In contrast to feature concatenation and traditional attention mechanisms, the dual-view MHSA method facilitates distinct and targeted processing of both contextual and sequential features, with each stream fine-tuned through dedicated attention weights. By integrating these dual perspectives, the method can synthesize insights from both views to foster a comprehensive understanding of traffic dynamics. This capability is critical for prediction tasks in complex environments where both contextual influences and historical patterns significantly impact outcomes.

Upon deriving the fused context-aware representations from the dual-view MHSA, they are subsequently integrated into a graph neural network to facilitate traffic speed forecasting, i.e., CKG-GNN. Specifically, we leverage traffic speed data alongside the context-aware representations from the preceding $a$ time steps to forecast traffic speed for the forthcoming $b$ time steps. In this study, we employ the diffusion convolution recurrent neural network (DCRNN) as our backbone GNN model, which models traffic forecasting as a diffusion process on a graph to capture both spatial and temporal dependencies through a sequence-to-sequence architecture, as detailed by \cite{li2017diffusion}. 
When integrating the CKG framework with DCRNN, the inputs comprise two components: raw traffic speed data from the past $a$ time steps, denoted as $(v_{t-a+1}, \dots, v_{t-1}, v_{t})$, and context-aware representations derived from the CKG for the same time window, denoted as $(\textbf{e}'_{fuse(t-a+1)}, \dots, \textbf{e}'_{fuse(t-1)}, \textbf{e}'_{fuse(t)})$.
The output is the predicted traffic speeds for the next $b$ time steps, denoted as $(v'_{t+1}, \dots, v'_{t+b})$.
During training, the objective is to minimize the mean absolute error between the predicted traffic speed $(v'_{t+1}, \dots, v'_{t+b})$ and actual traffic speeds $(v_{t+1}, \dots, v_{t+b})$ over future time steps, i.e.,
\begin{equation}
    \min \left( \frac{1}{b} \sum_{i=1}^{b} \left| v'_{t+i} - v_{t+i} \right| \right).
\end{equation}
The training is performed over 500 epochs with a batch size of 16, with the dataset divided into training, validation, and testing sets at ratios of 70\%, 10\%, and 20\%, respectively. The optimization is conducted using an Adam optimizer with an initial learning rate of 0.001, which is adjusted by a MultiStepLR scheduler that reduces the learning rate by 50\% at epochs 150, 250, 350, and 450 to enhance convergence. 
For the dual-view MHSA, we set the head numbers for the context view as 10 and for the sequence view as 4, tailored to the feature dimensions of each view.
Furthermore, the design of our CKG-GNN model supports real-time updates to spatial and temporal contexts by exclusively changing their attributes, without altering the CKG’s structure and the embedding process. These attribute changes are directly incorporated into the inputs of $(\textbf{e}'_{fuse(t-a+1)}, \dots, \textbf{e}'_{fuse(t-1)}, \textbf{e}'_{fuse(t)})$, which in turn impact the predicted traffic speeds.

\section{Experiments and Result}
\subsection{Datasets and processing}
The datasets used in this study encompass three categories: traffic speed data, spatial context data, and temporal context data. All datasets used in this study are publicly accessible. Detailed information about each dataset, including their respective time frames, resolutions, and data sources, is summarized in Table \ref{tab:dataset} in the Appendix.

\begin{figure}[htbp!]
  \centering
  \includegraphics[width=0.95\linewidth]{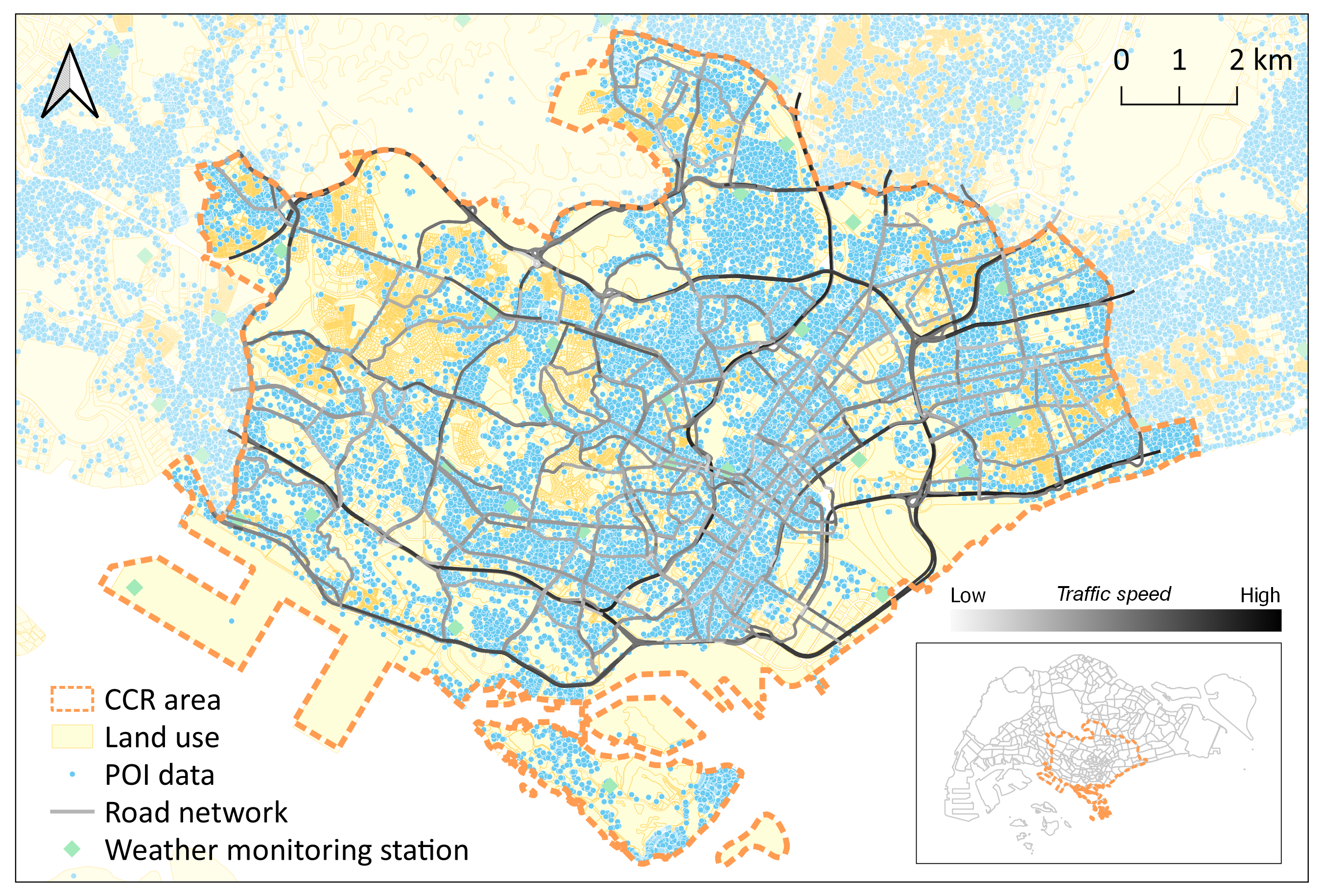}
  \caption{Overview of Singapore's Core Central Region (CCR) and the used datasets. The speed shown in the figure is a demo dataset collected from HERE technologies.}
  \label{fig:ccr_dataset}
\end{figure}

We sourced the traffic speed dataset from HERE technologies, focusing on Singapore's Core Central Region (CCR), i.e., the downtown area, due to its significant correlation between contextual information and traffic flow \cite{zhang2022street}. To investigate the performance of CKG-GNN in different scenarios, we construct three traffic speed datasets, i.e., (1) $V_{raw}$, 06:00-12:00 for about three-week workdays from 08/03/2022 to 25/03/2022, emphasizing the workday traffic patterns that specifically cover the morning peak hours; (2) $V_{work}$, 00:00-23:59 on Wednesday, 23/03/2022, a workday dataset; (3) $V_{rest}$, 00:00-23:59 on Saturday, 19/03/2022, a weekend dataset. 
The traffic speed datasets were collected every 2 minutes from HERE technologies. To enhance data quality and reduce noise, we aggregated the speed datasets into 10-min intervals by averaging. 
A demo of raw speed data in CCR is shown in Figure~\ref{fig:ccr_dataset}.

To establish $\mathbb{K}_S$, we collected three spatial context datasets, including road segments, POIs, and land uses shown in Figure~\ref{fig:ccr_dataset}.
(1) Road segments: The transportation network in CCR is built using shape information from HERE technologies, resulting in 1,606 viable road segments after correcting topological inaccuracies. The dataset also provides road segment lengths and free-flow speed information for creating $\mathbb{K}_S(road)$. 
(2) POIs: The POI dataset aggregates data sources from Singapore OneMap, DataMall, and OpenStreetMap, yielding 17 categories with 85,647 points across Singapore \cite{zhang2023incorporating}. It encompasses diverse classes, such as community, education, commercial, and residential areas, to form $\mathbb{K}_S(poi)$.
(3) Land uses: This dataset sourced from Singapore's Urban Redevelopment Authority in 2019, includes 28 reclassified categories such as business, residential, and park areas, offering a comprehensive view of the urban landscape to construct $\mathbb{K}_S(land)$.

Regarding the construction of $\mathbb{K}_T$, temporal contexts like time indicators, traffic jam factors, and weather conditions are collected due to their influential impact on traffic dynamics.
(1) Time: This context indicates the hour of the day and the day of the week to distinguish the temporal changes of traffic dynamics in composing $\mathbb{K}_T(time)$.
(2) Traffic jam factor: Collected every 2 minutes from HERE Technologies, this dataset provides real-time congestion levels ranging from 0.0 (free flow) to 10.0 (road closure) for each road segment. We aggregated these values into 10-min intervals to align with the temporal resolution of the traffic speed dataset, and then constructed $\mathbb{K}_T(jam)$.
(3) Weather conditions: We collected real-time data on air temperature, rainfall, and wind speed from Singapore's weather stations to build $\mathbb{K}_T(weather)$, aligning their temporal granularity with the traffic speed dataset by resampling them to 10-min intervals. Weather data was spatially associated with road segments based on proximity to the nearest weather stations. The distribution of weather monitoring stations is illustrated in Figure~\ref{fig:ccr_dataset}, and the types of weather conditions collected vary by station, with some stations capturing all three conditions while others collect only one or two, depending on the stations' settings.

\subsection{Spatial KG embedding evaluation}\label{sec:spat_kgeval}
Embedding $\mathbb{K}_S$ involves estimating MR's performance with different KGE models and parameters to select the optimal configuration by capturing the complex interactions of spatial factors.
In Table \ref{tab:spat_embed_buffer}, distance-based methods generally excel in embedding the spatial unit $\mathbb{K}_S$, as indicated by their lower both-side MR values compared to similarity-based methods. However, ComplEx, a similarity-based method, defies this trend by achieving the lowest MR (12.46 for Buffer[10-100] and Link[6]), suggesting its effectiveness in capturing spatial relationships in $\mathbb{K}_S$. Exclusively under the conditions of Buffer[10-100] and Link[-], TransE attains the best MR of 50.64.
The superior performance of ComplEx is attributed to its ability to model asymmetric relationships through complex-valued embeddings in $\mathbb{K}_S$ \cite{cao2024knowledge}. 
See Table \ref{tab:spat_embed_buffer_apd} in the Appendix for the evaluation results of MRR, H@5, and AH@K, which also demonstrate the best performance of ComplEx with Buffer[10-100] and Link[6].
Also, the result aligns with the notion that similarity-based methods like RESCAL and NTN require a substantial amount of facts for training to perform optimally.

\begin{table}[htbp!]
\centering
\caption{KG embedding performance (both-side MR) of the spatial unit $\mathbb{K}_S$ with different configurations of buffer [Dist]. There are two spatial links: Link[-] (no link) and Link[6] ([Order]=6).}
\label{tab:spat_embed_buffer}
\renewcommand\arraystretch{1.0}
\begin{tabular}{ccccccc}
\hline
\multirow{2}{*}{Model} & \multicolumn{2}{c}{Buffer{[}10-100{]}} & \multicolumn{2}{c}{Buffer{[}100-500{]}} & \multicolumn{2}{c}{Buffer{[}10-500{]}} \\ \cline{2-7} 
                       & Link{[}-{]}        & Link{[}6{]}       & Link{[}-{]}        & Link{[}6{]}        & Link{[}-{]}        & Link{[}6{]}       \\ \hline
TransE  & \textbf{50.64} & 20.40          & 72.89          & 33.25          & 70.27          & 39.44          \\ 
TransR  & 58.74          & 18.23          & 71.73          & 32.28          & 66.88          & 35.60          \\ 
KG2E    & 54.09          & 21.18          & 76.38          & 35.11          & 72.11          & 37.77          \\ \hline
RESCAL  & 764.17         & 824.76         & 705.22         & 793.25         & 720.09         & 785.08         \\ 
ComplEx & 53.66          & \textbf{12.46} & \textbf{66.52} & \textbf{24.86} & \textbf{57.99} & \textbf{27.88} \\ 
NTN     & 429.00         & 51.28          & 180.69         & 55.80          & 177.33         & 67.01          \\ \hline
\end{tabular}
\end{table}

In addition, Table \ref{tab:spat_embed_buffer} also delves into the impact of buffer distances on KG embeddings for $\mathbb{K}_S(poi)$ and $\mathbb{K}_S(land)$. Different buffer ranges, namely Buffer[10-100] and Buffer[100-500], are examined alongside a combined range of Buffer[10-500]. 
We find that Buffer[10-100] achieves the best performance for well-performing methods like TransE, TransR, KG2E, and ComplEx, suggesting that a closer proximity range is sufficient for associating POIs and land uses with the road network for effective embedding. Extending the buffer to include more distant POIs and land parcels may introduce noise, potentially confounding the embedding process.
Also, the MR evaluation under two spatial link conditions, Link[-] and Link[6], justifies the observations regarding KGE methods and buffer distance efficacy, underscoring the robustness of the embedding strategies across different spatial link scenarios.

\begin{table}[htbp!]
\centering
\caption{Both-side MR evaluation of various spatial links in the KG embedding of $\mathbb{K}_S$. Link[-] is no spatial link, and Link[1,...,6] refers to spatial links with [Order]=1,...,6.}
\label{tab:spat_embed_link}
\renewcommand\arraystretch{1.0}
\begin{tabular}{p{0.8cm}p{0.8cm}p{0.5cm}p{0.5cm}p{0.5cm}p{0.5cm}p{0.5cm}p{0.5cm}p{0.5cm}} 
\hline
Model   & Buffer                               & {[}-{]}         & {[}1{]}         & {[}2{]}         & {[}3{]}         & {[}4{]}         & {[}5{]}         & {[}6{]}         \\ \hline
ComplEx & \multirow{2}{*}{{[}10-100{]}}  & 53.66          & 49.38          & \textbf{41.72} & \textbf{33.26} & \textbf{23.21} & \textbf{17.09} & \textbf{12.46} \\  
TransE  &                                      & \textbf{50.64} & \textbf{48.10} & 43.70          & 37.94          & 30.81          & 24.44          & 20.40          \\ \hline
ComplEx & \multirow{2}{*}{{[}100-500{]}} & \textbf{66.52} & \textbf{63.36} & \textbf{58.60} & \textbf{50.69} & \textbf{41.04} & \textbf{31.90} & \textbf{24.86} \\ 
TransE  &                                      & 72.89          & 69.90          & 65.62          & 58.34          & 49.07          & 40.36          & 33.25          \\ \hline
ComplEx & \multirow{2}{*}{{[}10-500{]}}  & \textbf{57.99} & \textbf{57.26} & \textbf{53.97} & \textbf{48.52} & \textbf{42.19} & \textbf{34.78} & \textbf{27.88} \\  
TransE  &                                      & 70.27          & 68.58          & 64.82          & 60.65          & 55.01          & 47.47          & 39.44          \\ \hline
\end{tabular}
\end{table}

Given that both ComplEx and TransE demonstrate the potential to achieve the lowest MR in some scenarios in Table \ref{tab:spat_embed_buffer}, we further explore their performance across diverse spatial link configurations.
As shown in Table \ref{tab:spat_embed_link}, While TransE outperforms in the specific scenario of Link[-] and Link[1] for Buffer[10-100], ComplEx demonstrates superior performance across all other configurations.
A closer examination of the impact of different spatial links reveals a consistent trend: an enhancement in spatial links, as indicated by lower both-side MR values, correlates with improved embedding performance. This trend suggests that denser connections within the transportation network contribute positively to embedding $\mathbb{K}_S$, a pattern that holds across all buffer distances.
Table \ref{tab:spat_embed_link_apd} in the Appendix also corroborates the observation that ComplEx demonstrates superior performance across most configurations, with improved outcomes when additional spatial links are incorporated, as evidenced by the metrics of MRR, H@5, and AH@K.
Consequently, the optimal CKG configuration for embedding $\mathbb{K}_S$ appears to be Link[6] with Buffer[10-100], where ComplEx achieves the lowest both-side MR, specifically 12.46, implying its potential in providing the most suitable embeddings for subsequent traffic forecasting tasks.

\subsection{Temporal KG embedding evaluation}\label{sec:temp_kgeval}
Similar to embedding $\mathbb{K}_S$, selecting appropriate methods and parameters to embed the temporal unit $\mathbb{K}_T$ is also crucial. 
Table \ref{tab:temp_embed_time} demonstrates the both-side MR results for embedding $\mathbb{K}_T$ using both distance-based and similarity-based methods. Notably, KG2E consistently outperforms other methods across different configurations with the lowest MR values, benefiting from its approach of representing entities and relations through multi-variate Gaussian distributions \cite{he2015learning}. This feature makes KG2E incorporate uncertainties into the embedding process, capturing the inherent variability and complexity of relations in $\mathbb{K}_T$, especially in dynamic systems like traffic networks.
Table \ref{tab:temp_embed_time_apd} in the Appendix presents additional evaluation metrics: MRR, H@5, and AH@K, which suggest the selection of KG2E to ensure reliable and robust performance.
Regarding the impact of [PastMins], a distinct pattern emerges: larger [PastMins] values correlate with lower MR values across all methods, suggesting that a broader time buffer enhances embedding effectiveness. The cutoff at Past[60] is strategically chosen to align with the hourly link, and extending the time buffer beyond this point could introduce redundant complexity. Therefore, Past[60] is determined to be the optimal choice.
In addition, the above observations hold across two kinds of temporal links (Link[-] and Link[HDW]), reinforcing the findings' consistency and reliability.

\begin{table*}[htbp!]
\centering
\caption{KG embedding performance (both-side MR) of the temporal unit $\mathbb{K}_T$ with different configurations of [PastMins]. There are two temporal link situations, i.e., [-] referring to no links and [HDW] including hourly, daily, and weekly links.}
\label{tab:temp_embed_time}
\renewcommand\arraystretch{1.0}
\makebox[\textwidth]{
\begin{tabular}{ccccccccccccc}
\hline
\multirow{2}{*}{Model} & \multicolumn{2}{c}{Past{[}10{]}} & \multicolumn{2}{c}{Past{[}20{]}} & \multicolumn{2}{c}{Past{[}30{]}} & \multicolumn{2}{c}{Past{[}40{]}} & \multicolumn{2}{c}{Past{[}50{]}} & \multicolumn{2}{c}{Past{[}60{]}} \\ \cline{2-13} 
                       & {[}-{]}         & {[}HDW{]}      & {[}-{]}         & {[}HDW{]}      & {[}-{]}         & {[}HDW{]}      & {[}-{]}         & {[}HDW{]}      & {[}-{]}         & {[}HDW{]}      & {[}-{]}         & {[}HDW{]}      \\ \hline
TransE                 & 6.35            & 3.97           & 5.69            & 3.98           & 4.98            & 3.73           & 3.97            & 3.79           & 3.98            & 3.33           & 3.73            & 3.30           \\
TransR                 & 18.93           & 10.69          & 18.56           & 12.90          & 11.00           & 13.15          & 10.69           & 9.08              & 12.90           & 8.49              & 13.15           & 7.02           \\
KG2E                   & \textbf{5.47}   & \textbf{3.49}  & \textbf{4.13}   & \textbf{3.18}  & \textbf{3.75}   & \textbf{3.18}  & \textbf{3.49}   & \textbf{2.85}  & \textbf{3.18}   & \textbf{2.59}  & \textbf{3.18}   & \textbf{2.36}  \\ \hline
RESCAL                 & 282.31          & 406.15         & 407.05          & 405.33         & 407.25          & 407.00         & 406.15          & 405.54         & 405.33          & 406.28         & 407.00          & 407.20         \\
ComplEx                & 388.57          & 15.49          & 193.17          & 7.79           & 50.09           & 5.53           & 20.22           & 4.36           & 8.09            & 4.18           & 6.07            & 4.51           \\
NTN                    & 352.73          & 290.40         & 349.04          & 242.79         & 291.22          & 225.85         & 290.40          & 186.63         & 242.79          & 149.17         & 225.85          & 123.72         \\ \hline
\end{tabular}
}
\end{table*}

To further investigate the most effective temporal link configuration, Table \ref{tab:temp_embed_link} evaluates the both-side MR values for different temporal links with all [PastMins] using KG2E. It is noted that Link[HDW] consistently achieves the best performance, with the lowest both-side MR of 2.36 at Past[60].
Interestingly, the MR values for single temporal links—[hour], [day], and [week]—are identical. This uniformity suggests that when embedding $\mathbb{K}_T$, the temporal links alone (hourly, daily, or weekly) do not differentiate connections among road segments as they all evolve over time. However, when integrating all temporal links (Link[HDW]), $\mathbb{K}_T$ achieves an optimal structure for embedding.
Refer to Table \ref{tab:temp_embed_link_apd} in the Appendix for additional evaluation metric results of MRR, H@5, and AH@K, which also highlight the optimal performance of the Link[HDW] and Past[60] configuration for embedding $\mathbb{K}_T$.
In summary, embedding $\mathbb{K}_T$ using KG2E is most effective with Link[HDW] and Past[60], offering the most effective CKG embedding for the subsequent traffic forecasting tasks.

\begin{table}[htbp!]
\centering
\caption{Both-side MR evaluation of various temporal links in the KG embedding of $\mathbb{K}_T$ using KG2E. }
\label{tab:temp_embed_link}
\renewcommand\arraystretch{1.0}
\begin{tabular}{p{0.6cm}p{0.7cm}p{0.7cm}p{0.7cm}p{0.7cm}p{0.7cm}p{0.7cm}p{0.7cm}}
\hline
Model         & Link                   & Past{[}10{]}          & Past{[}20{]}        & Past{[}30{]}         & Past{[}40{]}       & Past{[}50{]}        & Past{[}60{]}  \\ \hline
KG2E          & {[}-{]}            & 5.47          & 4.13          & 3.75          & 3.49          & 3.18          & 3.18          \\ \hline
KG2E          & {[}hour{]}         & 4.13          & 3.75          & 3.49          & 3.18          & 3.18          & 2.85          \\ \hline
KG2E          & {[}day{]}          & 4.13          & 3.75          & 3.49          & 3.18          & 3.18          & 2.85          \\ \hline
KG2E          & {[}week{]}         & 4.13          & 3.75          & 3.49          & 3.18          & 3.18          & 2.85          \\ \hline
KG2E & {[}HDW{]} & \textbf{3.49} & \textbf{3.18} & \textbf{3.18} & \textbf{2.85} & \textbf{2.59} & \textbf{2.36} \\ \hline
\end{tabular}
\end{table}

\subsection{Performance of traffic speed prediction}
To investigate the performance of CKG-GNN models in traffic forecasting, we utilized speed datasets $V_{raw}$ for experiments. The results, as shown in Table \ref{tab:kggcn_alllink}, illustrate the evaluation metrics of various forecasting models over time horizons of 10, 60, and 120 minutes. 
We evaluate the performance of CKG-GNN models from three perspectives. First, we explore the impact of different CKG units on traffic prediction by developing three CKG-GNN variants, each integrating distinct context datasets: CKG-$\mathbb{K}_S$ (spatial only), CKG-$\mathbb{K}_T$ (temporal only), and CKG-$\mathbb{K}_{ST}$ (spatial and temporal).
Second, to benchmark CKG-GNN models against baseline GNN models, we utilize TGCN, STGCN, and DCRNN from the LibCity library to predict traffic speeds over identical time horizons \cite{wang2021libcity}. 
Third, for a stepwise comparison with other context-aware models using these context datasets, we investigate the performance of a multimodal context-based graph convolutional neural network (MCGCN) across three variants \cite{zhang2023incorporating}: MCGCN$_S$ (spatial only), MCGCN$_T$ (temporal only), and MCGCN$_{ST}$ (spatial and temporal). Each MCGCN model employs a DCRNN backbone with attention dimensions of 96 for both spatial and temporal contexts.
Each model is run five times to ensure a robust comparison, and the results are reported as the average with standard deviations (ave±std). The employed evaluation metrics include MAE (mean absolute error), quantifying the average magnitude of prediction errors; MAPE (mean absolute percentage error), which expresses these errors as a percentage, providing a scale-independent measure of accuracy but being sensitive to small denominators; and RMSE (root mean square error), which emphasizes larger errors by squaring the prediction differences before averaging, thereby providing sensitivity to larger error magnitudes.


\begin{table*}[htbp!]
\centering
\caption{Performance comparison of traffic speed prediction across models at 10, 60, and 120 minutes, and averaged over 10 to 120 minutes, using MAE (km/h), MAPE (\%), \textcolor{mycolor}{and RMSE (km/h)}. Each experiment was run five times to obtain its mean and standard deviation.}
\label{tab:kggcn_alllink}
\renewcommand\arraystretch{1.0}
\begin{tabular}{p{1.3cm}p{0.9cm}p{1.1cm}p{0.9cm}p{0.9cm}p{1.1cm}p{0.9cm}p{0.9cm}p{1.1cm}p{0.9cm}p{0.9cm}p{1.1cm}p{0.9cm}}
\hline
\multicolumn{1}{c}{\multirow{2}{*}{Model}} & \multicolumn{3}{c}{10 min}                                                    & \multicolumn{3}{c}{60 min}                                                    & \multicolumn{3}{c}{120 min}                                                   & \multicolumn{3}{c}{Ave (10-120 min)}                                          \\ \cline{2-13} 
\multicolumn{1}{c}{}                       & \multicolumn{1}{c}{MAE} & \multicolumn{1}{c}{MAPE} & \multicolumn{1}{c}{\textcolor{mycolor}{RMSE}} & \multicolumn{1}{c}{MAE} & \multicolumn{1}{c}{MAPE} & \multicolumn{1}{c}{\textcolor{mycolor}{RMSE}} & \multicolumn{1}{c}{MAE} & \multicolumn{1}{c}{MAPE} & \multicolumn{1}{c}{\textcolor{mycolor}{RMSE}} & \multicolumn{1}{c}{MAE} & \multicolumn{1}{c}{MAPE} & \multicolumn{1}{c}{\textcolor{mycolor}{RMSE}} \\ \hline
TGCN                                       & 4.82±0.04               & 20.66±0.30               & \textcolor{mycolor}{6.41±0.05}                & 5.34±0.03               & 24.10±0.17               & \textcolor{mycolor}{7.21±0.04}                & 5.52±0.06               & 25.34±0.31               & \textcolor{mycolor}{7.44±0.06}                & 5.32±0.03               & 23.92±0.15               & \textcolor{mycolor}{7.16±0.04}                \\
STGCN                                      & 3.29±0.05               & 13.93±0.12               & \textcolor{mycolor}{4.73±0.09}                & 3.78±0.05               & 16.19±0.44               & \textcolor{mycolor}{5.48±0.05}                & 3.76±0.08               & 16.21±0.61               & \textcolor{mycolor}{5.42±0.07}                & 3.72±0.04               & 15.95±0.37               & \textcolor{mycolor}{5.38±0.03}                \\
DCRNN                                      & 3.14±0.00               & 13.15±0.02               & \textcolor{mycolor}{4.50±0.01}                & 3.69±0.01               & 16.13±0.07               & \textcolor{mycolor}{5.37±0.01}                & 3.73±0.01               & 16.47±0.13               & \textcolor{mycolor}{5.40±0.01}                & 3.64±0.01               & 15.86±0.07               & \textcolor{mycolor}{5.27±0.01}                \\ \hline
\textcolor{mycolor}{MCGCN$_S$}                                  & \textcolor{mycolor}{3.18±0.06}               & \textcolor{mycolor}{13.19±0.14}               & \textcolor{mycolor}{4.49±0.03}                & \textcolor{mycolor}{3.62±0.05}               & \textcolor{mycolor}{15.77±0.23}               & \textcolor{mycolor}{5.29±0.05}                & \textcolor{mycolor}{3.66±0.07}               & \textcolor{mycolor}{16.10±0.36}               & \textcolor{mycolor}{5.34±0.08}                & \textcolor{mycolor}{3.58±0.06}      & \textcolor{mycolor}{15.54±0.25}               & \textcolor{mycolor}{{5.20±0.05}}       \\
\textcolor{mycolor}{MCGCN$_T$}                                  & \textcolor{mycolor}{3.20±0.10}               & \textcolor{mycolor}{13.29±0.15}               & \textcolor{mycolor}{4.50±0.08}                & \textcolor{mycolor}{3.58±0.03}               & \textcolor{mycolor}{15.42±0.10}               & \textcolor{mycolor}{5.23±0.03}                & \textcolor{mycolor}{3.60±0.01}               & \textcolor{mycolor}{15.58±0.15}               & \textcolor{mycolor}{5.26±0.03}                & \textcolor{mycolor}{3.55±0.03}               & \textcolor{mycolor}{15.24±0.06}               & \textcolor{mycolor}{5.15±0.02}                \\
\textcolor{mycolor}{MCGCN$_{ST}$}                                 & \textcolor{mycolor}{3.17±0.07}               & \textcolor{mycolor}{13.22±0.07}               & \textcolor{mycolor}{4.47±0.04}                & \textcolor{mycolor}{3.54±0.03}               & \textcolor{mycolor}{15.29±0.08}               & \textcolor{mycolor}{5.19±0.04}                & \textcolor{mycolor}{3.57±0.03}               & \textcolor{mycolor}{15.45±0.09}               & \textcolor{mycolor}{5.22±0.04}                & \textcolor{mycolor}{3.51±0.04}               & \textcolor{mycolor}{15.10±0.08}               & \textcolor{mycolor}{5.11±0.03}                \\ \hline
CKG-$\mathbb{K}_S$                         & 3.16±0.00               & 13.12±0.02               & \textcolor{mycolor}{4.63±0.01}                & 3.65±0.03               & 15.61±0.11               & \textcolor{mycolor}{5.44±0.05}                & 3.64±0.03               & 15.94±0.16               & \textcolor{mycolor}{5.40±0.04}                & 3.60±0.02               & {15.46±0.09}      & \textcolor{mycolor}{5.35±0.03}                \\
CKG-$\mathbb{K}_T$                         & 3.09±0.01               & 12.95±0.05               & \textcolor{mycolor}{4.42±0.00}                & 3.56±0.01               & 15.37±0.09               & \textcolor{mycolor}{5.23±0.02}                & 3.56±0.02               & 15.39±0.10               & \textcolor{mycolor}{5.20±0.02}                & {3.51±0.01}      & {15.10±0.08}      & \textcolor{mycolor}{{5.13±0.01}}       \\
CKG-$\mathbb{K}_{ST}$                      & 3.08±0.01               & 12.88±0.05               & \textcolor{mycolor}{4.42±0.01}                & 3.52±0.00               & 15.04±0.15               & \textcolor{mycolor}{5.19±0.01}                & 3.47±0.02               & 14.86±0.11               & \textcolor{mycolor}{5.10±0.02}                & \textbf{3.46±0.01}      & \textbf{14.76±0.09}      & \textcolor{mycolor}{\textbf{5.08±0.01}}       \\ \hline
\end{tabular}
\end{table*}

The results in Table \ref{tab:kggcn_alllink} underscore the significant boost in forecasting performance achieved by integrating the CKG with GNN across various time horizons.  
When examining the average performance from 10 to 120 minutes, DCRNN stands out among the baseline models with an MAE of 3.64±0.01, a MAPE of 15.86±0.07\%, and an RMSE of 5.27±0.01. 
Using DCRNN as the backbone, both context-aware models, i.e., MCGCN and CKG-GNN, demonstrate enhanced performance when incorporating contextual information into traffic speed predictions. However, CKG-GNN consistently outperforms MCGCN when both spatial and temporal contexts are integrated. In detail, CKG-$\mathbb{K}_{ST}$ records an MAE of 3.46±0.01, a MAPE of 14.76±0.09\%, and an RMSE of 5.08±0.01, marking improvements over DCRNN of 0.18 for MAE, 1.10\% for MAPE, and 0.19 for RMSE. Compared to MCGCN$_{ST}$, the improvements are 0.05 for MAE, 0.34\% for MAPE, and 0.03 for RMSE. These results demonstrate the superiority of CKG-GNN models, illustrating that they not only surpass baseline GNN models in performance but also excel beyond other advanced context-aware models. 
A step-by-step comparison between MCGCN and CKG-GNN reveals that MCGCN$_{ST}$ performs suboptimally when both contexts are incorporated; however, MCGCN$_{S}$, which considers only the spatial context, exhibits better MAE and RMSE than CKG-$\mathbb{K}_{S}$, although CKG-$\mathbb{K}_{S}$ achieves a better MAPE. This suggests that MCGCN's hierarchical learning effectively captures the spatial dependencies inherent in the context. In the configuration that considers only the temporal context, CKG-$\mathbb{K}_{T}$ shows an advantage over MCGCN$_{S}$, improving MAE, MAPE, and RMSE by 0.04, 0.14\%, and 0.02, respectively.
Overall, these findings affirm that using knowledge graphs to structure context datasets enhances the capture of spatial and temporal dependencies, thereby improving traffic prediction accuracy. 
Different from existing context-aware traffic forecasting methods such as MCGCN, which necessitate distinct models to accommodate the characteristics of spatial and temporal contexts \cite{zhang2023incorporating}, the proposed CKG-GNN model ensures uniformity by utilizing a consistent knowledge graph format to model both context types. Moreover, this approach not only streamlines the integration of additional context datasets without the need to develop specialized models for learning their representations but also demonstrates superior performance.

Moreover, it is observed that integrating the temporal context into DCRNN generally yields greater improvements than incorporating the spatial context. This is evidenced by CKG-$\mathbb{K}_{T}$ outperforming CKG-$\mathbb{K}_{S}$ and MCGCN$_{T}$ surpassing MCGCN$_{S}$, suggesting that temporal factors have a more pronounced impact on forecasting accuracy than spatial factors.
This improvement is attributed to the direct impact of temporal factors on traffic forecasting due to their alignment with the dynamic nature of traffic flow. In contrast, spatial contexts contribute by delineating the intricate supply-demand relationship of human mobility over urban spaces, a complex aspect challenging to leverage effectively in traffic prediction models \cite{zhang2023incorporating}.

The individual performance of 10, 60, and 120 minutes reveals that the CKG-enhanced models consistently outperform the baselines, with a minor exception at the 10-min prediction using only $\mathbb{K}_S$. Although the 10-min MAE and RMSE for $\mathbb{K}_S$ is slightly worse than DCRNN, the MAPE still shows an improvement.
Furthermore, CKG-$\mathbb{K}_{ST}$ outperforms all the other models from 10-min to 120-min prediction, suggesting that the combination of spatio-temporal contexts contributes to the predictive reliability over both short and longer-term horizons.
Regarding error progression from 10-min to 120-min forecasts, baseline models like DCRNN exhibit a gradual increase in MAE, but this error accumulation can be mitigated when context-aware knowledge graphs are integrated. Despite an increase in MAE from 10-min to 60-min forecasts across $\mathbb{K}_S$, $\mathbb{K}_T$, and $\mathbb{K}_{ST}$, their performance from 60 minutes to 120 minutes remains stable or even improves, such as the MAE decrease from 3.52±0.00 to 3.46±0.01 for CKG-$\mathbb{K}_{ST}$. This stability indicates the potent effect of CKG in enhancing longer-term traffic speed predictions.

\subsection{Dual-view feature importance}
\begin{figure}[htbp!]
  \centering
  \includegraphics[width=0.95\linewidth]{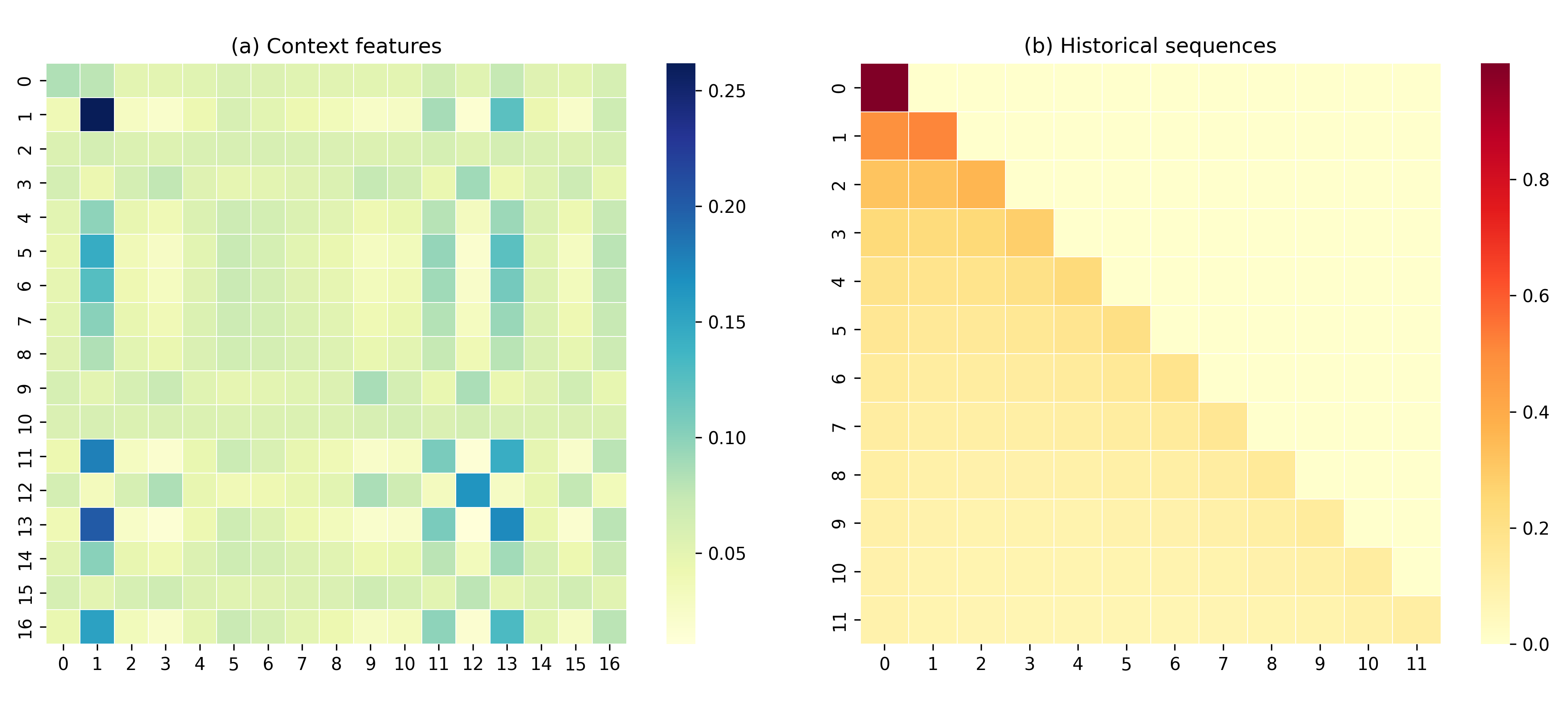}
  \caption{Attention weight heatmaps for context-view and sequence-view features in CKG-$\mathbb{K}_{ST}$. (a) Heatmap for context-view features with $[0-9]_a$ for the spatial unit and $[10-16]_a$ for the temporal unit. $[0]_a$: road-only spatial feature; $[1-3]_a$: relation-dependent features for roads, POIs, and land uses, respectively; $[4-9]_a$: spatial links from 1 to 6. $[10]_a$: road-only temporal feature; $[11-13]_a$: relation-dependent features for time indicators, jam factors, and weather conditions, respectively; $[14-16]_a$: temporal links for hour, day, and week, respectively. (b) Heatmap for sequence-view features in the last 12 time slots with masks. $[0]_b$: the earliest; $[11]_b$: the latest.}
  \label{fig:atten_weights}
\end{figure}

Figure~\ref{fig:atten_weights}(a) demonstrates the heatmap derived from context-view attention weights in Equation~\ref{eq:mhsa} to investigate the contribution of context features. 
For the spatial unit, the prominence of the relation-dependent road feature at $[1]_a$ stands out, surpassing the influence of the road-only spatial feature at $[0]_a$ and the road-only temporal feature at $[10]_a$. This phenomenon suggests that isolated road representations without additional contextual information are less predictive.
In examining the effectiveness of different spatial contexts, land uses at $[3]_a$ receive relatively higher importance scores compared to POIs at $[2]_a$. Despite this, both display less influence on traffic prediction compared to the relation-dependent road features at $[1]_a$. This lesser influence can be attributed to their indirect association with traffic systems, in contrast to the direct connections exhibited by road networks. The relatively high importance between land uses at $[3]_a$ and jam factors at $[12]_a$ underscores the impact of land use on traffic congestion factors, thereby contributing to traffic predictions.
Moreover, the weights in Rows $[4-8]_a$ are notably elevated, indicating that spatial links—ranging from proximate to extended connections—play a crucial role in capturing the spatial dependency affecting traffic prediction. These links also exhibit a strong interplay with temporal features, particularly time indicators at Col $[11]_a$ and weather conditions at Col $[13]_a$.

Regarding the temporal unit, all features within $[11-13]_a$ manifest as critical. First, time indicators and weather conditions in $[11,13]_a$ demonstrate substantial weights individually and mutually, indicating their strong influence on traffic speed predictions.
Their interaction with the relation-dependent road feature at Col $[1]_a$ is also of great importance.
Second, the jam factor at $[12]_a$ stands out for its significant self-weight, aligning with the intuitive understanding that the congestion level is directly predictive of traffic speed with an immediate impact. This emphasizes the critical role of jam factors as a key context feature in traffic forecasting.
Furthermore, hourly and weekly patterns in Rows $[14,16]_a$ carry high weights, whereas daily patterns in Row $[15]_a$ appear less influential. The prominence of hourly patterns can be attributed to their immediate relevance to the prediction timeframe. Meanwhile, weekly patterns demonstrate consistent regularity, capturing the cyclical nature of traffic flow where behaviors on a specific day, like Monday, tend to mirror those from the previous week. In contrast, daily patterns exhibit a weaker correlation, suggesting that the connection between consecutive days, such as Friday to the following Monday, possesses limited predictive value in this context.

On the other hand, Figure~\ref{fig:atten_weights}(b) provides an attention-weight heatmap from the sequence view over the last 12 time slots, where $[0]_b$ denotes the earliest and $[11]_b$ denotes the latest time slot. 
The upper-right corner of the heatmap is filled with zeros, a direct result of employing MHSA masks to prevent the model from accessing future information, ensuring that predictions are based solely on past and present data.
The heatmap illustrates that cells along and near the diagonal possess higher weights, underscoring the model's recognition of more immediate past events as more indicative of the current traffic speed. For instance, in Row $[1]_b$, the weight at Col $[1]_b$ is greater than that at Col $[0]_b$, aligning with the self-attention mechanism's design to assign more importance to proximate temporal information.
Moreover, the heatmap presents a gradient of diminishing weights progressing downward from Row $[0]_b$ to $[11]_b$, illustrating a gradual reduction in the influence of historical time slots on the present prediction. Overall, the heatmap thus not only visualizes the sequential dependencies within the time-series data but also reveals the diminishing impact of distant past information on current traffic speed forecasting.

\section{Discussion}
\subsection{CKG embedding analysis}\label{sec:discuss_kge}
\begin{figure}[htbp!]
  \centering
  \includegraphics[width=0.95\linewidth]{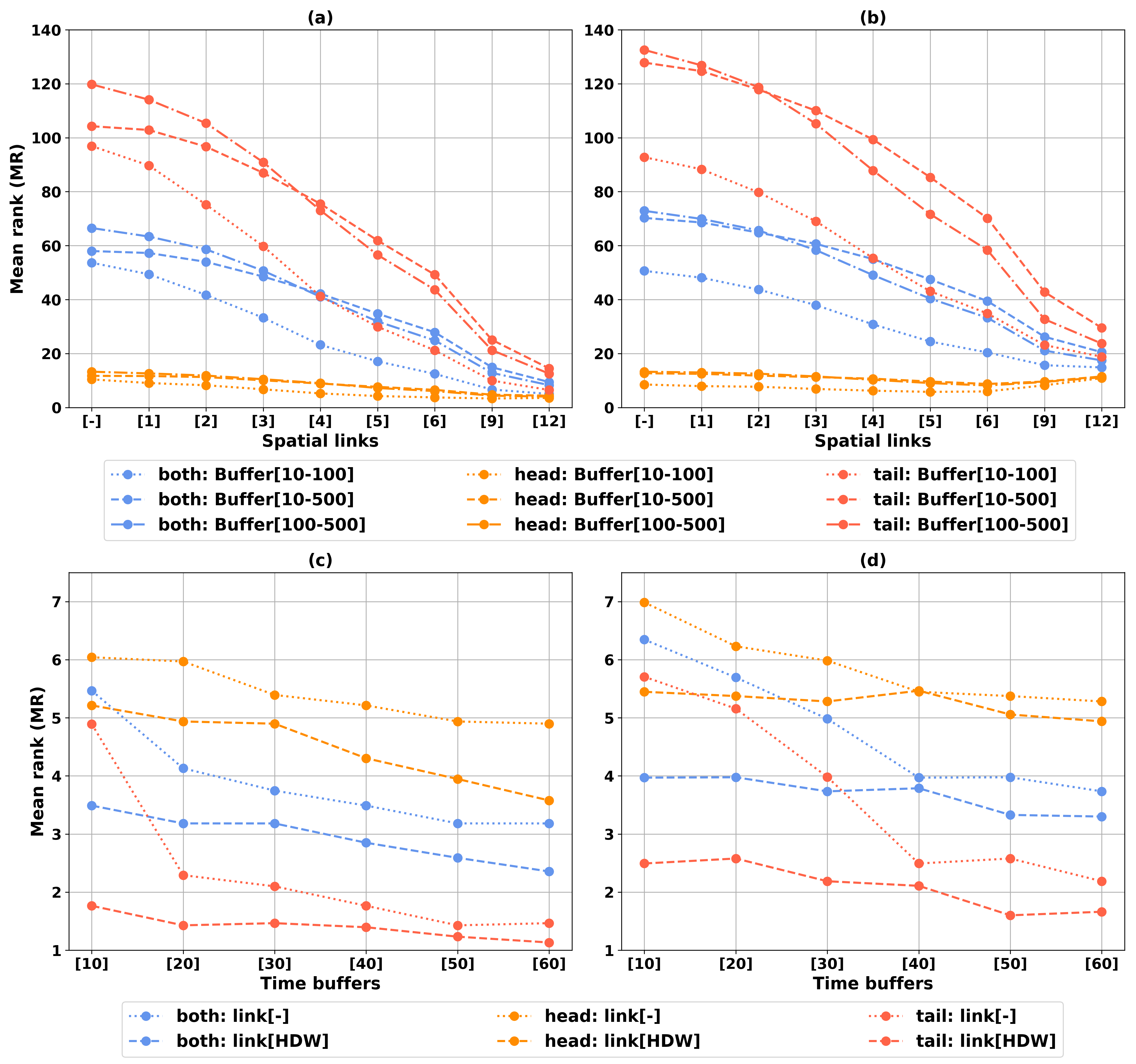}
  \caption{Embedding performance of both-side (b), left-side (head, h), and right-side (tail, t) link predictions for $\mathbb{K}_S$ and $\mathbb{K}_T$. (a) ComplEx for $\mathbb{K}_{S}$. (b) TransE for $\mathbb{K}_{S}$. (c) KG2E for $\mathbb{K}_{T}$. (d) TransE for $\mathbb{K}_{T}$.}
  \label{fig:ckg_embed}
\end{figure}

In Sections~\ref{sec:spat_kgeval} and~\ref{sec:temp_kgeval}, we have analyzed the both-side evaluation of CKG embeddings for $\mathbb{K}_{S}$ and $\mathbb{K}_{T}$. A further exploration into the left-side (head prediction) and right-side (tail prediction) evaluations can unveil additional insights, as shown in Figure~\ref{fig:ckg_embed}. In detail, Figure~\ref{fig:ckg_embed}(a, b) employs ComplEx and TransE methods across diverse spatial buffer ranges and spatial links, while Figure~\ref{fig:ckg_embed}(c, d) leverages KG2E and TransE methods to explore different time buffers and temporal links.

Regarding $\mathbb{K}_S$, we have several findings. 
(1) Head and tail predictions. An imbalance in MR values for left-side and right-side evaluations is observed for both methods in Figure~\ref{fig:ckg_embed}(a, b), with the both-side case presenting intermediate values. This imbalance reveals that predicting the head entity (e.g., $poiType$ and $landType$) given the tail entity and the relation is more straightforward compared to the reverse. We infer that the asymmetry stems from the more distinguishable spatial distribution of POIs and land types, making them easier to associate with specific road segments compared to predicting a road based on its surrounding urban functions.
(2) Spatial link influences. A decline in MR with increasing link numbers suggests that additional spatial links bolster the $\mathbb{K}_{S}$'s ability to delineate interactions and relationships between roads, POIs, and land uses. This trend is consistently observed across all buffer distances, indicating that denser connections within the transportation network can enhance the embedding performance of $\mathbb{K}_S$. To further explore the link influence, we also investigate the embedding performance for Link[9] and Link[12] by adding more spatial links. However, the inclusion of Link[9] and Link[12] does not yield a large performance enhancement beyond Link[6]. For TransE, the performance on the right-side prediction even deteriorates for Link[9] and Link[12]. Hence, Link[6] is recommended for an optimal parameter of $\mathbb{K}_{S}$, balancing performance gains and computational resource usage.
(3) Spatial buffer selection. We find that Buffer[10-100] achieves the best MR scores across three-side predictions using both models, corroborating findings from Section~\ref{sec:spat_kgeval}. This particular buffer range encapsulates essential spatial relationships in $\mathbb{K}_{S}$ among entities, striking a balance where the context is rich without being overly broad. 

Similarly, we analyze the three-side MR values for $\mathbb{K}_{T}$ using KG2E and TransE methods in Figure~\ref{fig:ckg_embed}(c, d).
(1) Head and tail predictions. We also observed the imbalanced phenomenon of left-side and right-side link predictions in embedding $\mathbb{K}_{T}$. For KG2E and TransE with different temporal links, predicting the head generally yields worse MR scores than predicting the tail. However, this imbalance becomes less evident compared to $\mathbb{K}_{S}$, which suggests a more uniform structure of the temporal unit.
(2) Temporal context dependency. The MR values demonstrate a consistent decrease as the time buffers increase from Past[10] to Past[60] for all cases of embedding $\mathbb{K}_{T}$. This pattern underscores the importance of including historical information to enhance link prediction accuracy in $\mathbb{K}_{T}$. Moreover, integrating more temporal links, i.e., Link[HDW], consistently bolsters MR scores for both models across all configurations, reinforcing the findings in Section~\ref{sec:temp_kgeval}. Both broadening the time buffer range from Past[10] to Past[60] and adding more temporal links from Link[-] to Link[HDW] introduce stronger temporal dependencies within $\mathbb{K}_{T}$, thereby improving the embedding effectiveness across various KGE techniques.

\subsection{Forecasting traffic speed in limited datasets}
\begin{figure}[htbp!]
  \centering
  \includegraphics[width=0.95\linewidth]{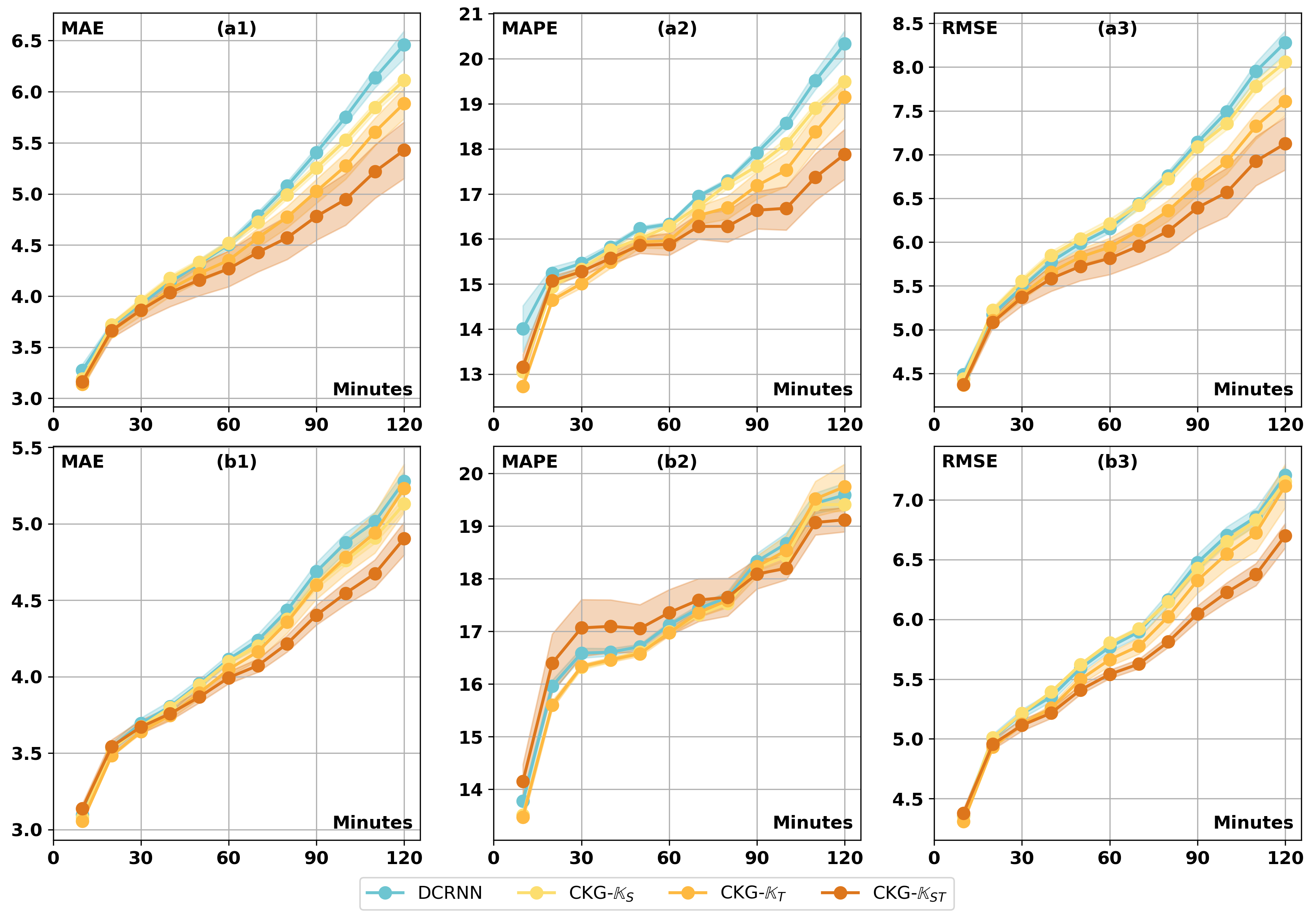}
  \caption{Performance evaluation (ave±std) of CKG-enhanced models and DCRNN for traffic speed prediction from 10 to 120 minutes. (a1-a3) MAE, MAPE, and RMSE for $V_{work}$. (b1-b3) MAE, MAPE, and RMSE for $V_{rest}$.}
  \label{fig:limited_metrics}
\end{figure}

To explore CKG-GNN's efficacy in scenarios with potentially limited data, we apply the CKG-GNN models to two small datasets and investigate their traffic prediction performance in Table~\ref{tab:kggcn_limited} and Figure~\ref{fig:limited_metrics}, i.e., one from a typical workday $V_{work}$ and the other from a rest day $V_{rest}$. We introduced the mean scaled interval score (MSIS) metric to quantify the uncertainty surrounding the point forecasts within a 95\% prediction interval, as shown in Table \ref{tab:kggcn_limited}. MSIS rewards models for providing narrow prediction intervals but penalizes them when actual values fall outside these intervals \cite{gneiting2007strictly}. A lower MSIS indicates a more reliable uncertainty estimate, which is crucial for providing enhanced insights into the model’s confidence in its predicted outcome \cite{lana2024measuring}. For details on MSIS computation, refer to Appendix\ref{app:msis}. The divergence in day types is expected to manifest itself in the model's performance due to the inherent differences in traffic patterns between workdays and rest days. The absence of weekly and daily temporal links in these single-day datasets leads to that only hourly links in $\mathbb{K}_{T}$ are utilized.

In Table~\ref{tab:kggcn_limited}, the CKG-enhanced models display superiority in 10-120 minute speed prediction for both workday and rest day according to the consistently lower average MAE, MAPE, and RMSE values compared to the baseline DCRNN model. 
This demonstrates that CKG can enhance the capacity of DCRNN to improve traffic forecasting accuracy, even with limited training data, through the integration of contextual information. 
The MSIS values also support this observation. CKG-$\mathbb{K}_{ST}$ achieved MSIS scores of 2.10 and 1.97, compared to 2.62 and 2.36 for DCRNN on the $V_{work}$ and $V_{rest}$ datasets, respectively, indicating lower uncertainty in the prediction intervals when integrating $\mathbb{K}_{ST}$.
Since $V_{work}$ and $V_{rest}$ represent distinct traffic patterns, the improved accuracy, reflected in lower MAE, MAPE, RMSE, and the reduced uncertainty, indicated by lower MSIS, demonstrate that incorporating context-aware representations enables the model to learn domain-specific information, resulting in more accurate traffic forecasts despite the differences in traffic patterns.

In addition, when comparing the two limited datasets using $\mathbb{K}_{S}$, $\mathbb{K}_{T}$, and their combination $\mathbb{K}_{ST}$, we find that CKG-$\mathbb{K}_{ST}$ generally performs best, followed by CKG-$\mathbb{K}_{T}$, and then CKG-$\mathbb{K}_{S}$ for MAE, RMSE, and MSIS. The only exception is the MAPE for $V_{rest}$, where CKG-$\mathbb{K}_{S}$ yields the lowest value. This further verifies the observation in Table \ref{tab:kggcn_alllink} that integrating temporal contexts into DCRNN typically leads to greater improvements than incorporating spatial contexts, even with limited datasets. Moreover, these results highlight the complementary strengths of spatial and temporal contexts. Spatial contexts enhance prediction accuracy by associating traffic systems with their surrounding environments, indirectly capturing human mobility patterns \cite{zhang2022street}. Conversely, temporal contexts directly capture the dynamic variations in traffic flow, making their influences more immediate \cite{zhang2023incorporating}. The integration of both contexts allows for simultaneous leveraging of their characteristics, as shown in Figure \ref{fig:atten_weights}, yielding superior results and validating the effectiveness of combining $\mathbb{K}_{S}$ and $\mathbb{K}_{T}$ into $\mathbb{K}_{ST}$.


\begin{table}[htbp!]
\centering
\caption{Performance evaluation of CKG-enhanced models and DCRNN for traffic speed prediction (10–120 minutes) on two limited datasets.}
\label{tab:kggcn_limited}
\renewcommand\arraystretch{1.0}
\begin{tabular}{clcccc}
\hline
\textcolor{mycolor}{Dataset}                         & \multicolumn{1}{c}{Model} & MAE                & MAPE                & \textcolor{mycolor}{RMSE}               & \textcolor{mycolor}{MSIS}          \\ \hline
\multirow{4}{*}{\textcolor{mycolor}{$V_{work}$}} & DCRNN                     & 4.79±0.05          & 16.97±0.12          & \textcolor{mycolor}{6.52±0.06}          & \textcolor{mycolor}{2.62}          \\
                            & CKG-$\mathbb{K}_S$        & 4.70±0.03          & 16.62±0.08          & \textcolor{mycolor}{6.48±0.05}          & \textcolor{mycolor}{2.49}          \\
                            & CKG-$\mathbb{K}_T$        & 4.54±0.09          & 16.27±0.21          & \textcolor{mycolor}{6.18±0.10}          & \textcolor{mycolor}{2.46}          \\
                            & CKG-$\mathbb{K}_{ST}$     & \textbf{4.38±0.17} & \textbf{16.00±0.24} & \textcolor{mycolor}{\textbf{5.97±0.20}} & \textcolor{mycolor}{\textbf{2.10}} \\ \hline
\multirow{4}{*}{\textcolor{mycolor}{$V_{rest}$}} & DCRNN                     & 4.23±0.04          & 17.32±0.09          & \textcolor{mycolor}{5.94±0.05}          & \textcolor{mycolor}{2.36}          \\
                            & CKG-$\mathbb{K}_S$        & 4.18±0.02          & \textbf{17.15±0.05} & \textcolor{mycolor}{5.93±0.02}          & \textcolor{mycolor}{2.25}          \\
                            & CKG-$\mathbb{K}_T$        & 4.16±0.06          & 17.20±0.12          & \textcolor{mycolor}{5.83±0.08}          & \textcolor{mycolor}{2.28}          \\
                            & CKG-$\mathbb{K}_{ST}$     & \textbf{4.07±0.04} & 17.40±0.36          & \textcolor{mycolor}{\textbf{5.65±0.04}} & \textcolor{mycolor}{\textbf{1.97}} \\ \hline
\end{tabular}
\end{table}

We then present the MAE, MAPE, and RMSE values for traffic prediction over the 10-120 minute range for a more detailed comparison in Figure \ref{fig:limited_metrics}. The central line represents the average performance, while the surrounding margin illustrates its standard deviation.
The performance metrics across different time horizons align with the findings in Table \ref{tab:kggcn_limited}, demonstrating that CKG-$\mathbb{K}_{ST}$ outperforms the other models in most cases. For $V_{work}$, the CKG-$\mathbb{K}_{ST}$ model consistently achieves the lowest MAE, MAPE, and RMSE values across the majority of time horizons in Figure \ref{fig:limited_metrics}(a1, a2, a3), while DCRNN produces higher errors throughout the 10-120 minute range. Notably, the reduction in MAE, MAPE, and RMSE between CKG-$\mathbb{K}_{ST}$ and DCRNN becomes more pronounced as the prediction horizon extends from 10 to 120 minutes, suggesting that the integration of CKG becomes especially effective for longer-term predictions.
Similarly, for $V_{rest}$, CKG-$\mathbb{K}_{ST}$ remains the best-performing model in terms of MAE and RMSE in Figure \ref{fig:limited_metrics}(b1, b3), apart from MAPE in Figure \ref{fig:limited_metrics}(b2). The trend of greater error reduction as the time horizons extend from 10 to 120 minutes is also observed in this dataset.

As previously noted, MAPE is an exception for $V_{rest}$, where the integration of $\mathbb{K}_{ST}$ proves less effective compared to its performance in MAE and RMSE.
Specifically, CKG-$\mathbb{K}_{ST}$ exhibits higher MAPE values during the 10-70 minute horizon but achieves lower values from 90 to 120 minutes. In contrast, CKG-$\mathbb{K}_{ST}$ outperforms other models in terms of MAE and RMSE across most time horizons. This discrepancy stems from the distinct ways MAE, MAPE, and RMSE handle errors: MAE treats all errors equally to measure overall bias, MAPE is highly sensitive to small actual speed values, and RMSE places more emphasis on larger errors.
Thus, the strong performance in MAE and RMSE highlights CKG-$\mathbb{K}_{ST}$'s ability to reduce overall and large errors. However, MAPE underperforms due to its sensitivity to low actual speed values, where small base speeds can result in disproportionately high percentage errors. This makes MAPE less reliable in this scenario. Moreover, the lack of cyclical patterns in the single-day dataset, such as daily and weekly trends, also complicates traffic speed prediction.

\subsection{Computational cost and practical implications}
Integrating context-aware knowledge graphs into the GNN model introduces additional parameters, thereby increasing the model's complexity. To illustrate this, we present the parameter numbers and computational costs for our CKG-GNN model compared to the baseline DCRNN model in Table \ref{tab:model_cost}. The CKG-GNN model includes 2291.7K parameters, significantly more than the 372.4K parameters of the DCRNN model.
However, the improvements in prediction accuracy justify these changes. For example, using the $V_{raw}$ dataset, the CKG-GNN model achieves reductions in MAE for 10-minute and 120-minute predictions by 0.06 and 0.26, respectively. Moreover, with limited datasets of $V_{work}$ and $V_{rest}$, CKG-GNN significantly enhances MAE by 1.03 in Figure \ref{fig:limited_metrics}(a1) and 0.38 in Figure \ref{fig:limited_metrics}(b1) for 120-minute predictions compared to DCRNN.
Regarding computational costs, benchmarking experiments were conducted on a machine equipped with a single RTX 4090 GPU and a 16-core CPU. We distinguish two main costs here. First, embeddings of $\mathbb{K}_S$ and $\mathbb{K}_T$ required 911.1s and 105.6s, respectively, performed only once without further computational overhead. Second, training times for CKG-GNN are longer than for DCRNN due to higher parameter counts, specifically 19485.8s compared to 3746.2s. This trade-off is mitigated by the minimal impact on inference times, with CKG-GNN requiring 7.5s versus 1.7s for DCRNN. Despite the slightly longer inference time for CKG-GNN, the duration is manageable and practical for real-world applications.


\begin{table}[htbp!]
\centering
\textcolor{mycolor}{
\caption{Parameter numbers and computation costs for DCRNN and CKG-GNN using $V_{raw}$. '-' indicates not applicable.}
\label{tab:model_cost}
\renewcommand\arraystretch{1.0}
\begin{tabular}{cc|cccc}
\hline
\multicolumn{2}{c|}{Method}                    & \begin{tabular}[c]{@{}c@{}}\#Params\\ (K)\end{tabular} & \begin{tabular}[c]{@{}c@{}}Embed\\ time (s)\end{tabular} & \begin{tabular}[c]{@{}c@{}}Train\\ time (s)\end{tabular} & \begin{tabular}[c]{@{}c@{}}Inference\\ time (s)\end{tabular} \\ \hline
\multicolumn{2}{c|}{DCRNN}                     & 372.4                                                  & -                                                        & 3746.2                                                   & 1.7                                                          \\ \hline
\multirow{3}{*}{CKG} & Embed $\mathbb{K}_{S}$ & 108.5                                                  & 911.1                                                    & -                                                        & -                                                            \\
                     & Embed $\mathbb{K}_{T}$ & 99.1                                                   & 105.6                                                    & -                                                        & -                                                            \\
                     & CKG-GNN                 & 2291.7                                                 & -                                                        & 19485.8                                                  & 7.5                                                          \\ \hline
\end{tabular}
}
\end{table}

The proposed method, despite its increased complexity, demonstrates strong generality through its integration with various publicly accessible context datasets, such as the free traffic API provided by HERE Technologies and the OSM platform. This accessibility facilitates the application of our framework to other urban settings. Furthermore, the framework is designed to incorporate additional types of contextual information, which could potentially enhance the CKG framework's predictive capabilities by including broader contextual variables. In this case, although new entities and relations would be integrated into the CKG, the core elements of our framework would continue to function effectively.
Moreover, the proposed CKG framework offers scalability through increased data volumes without affecting the CKG structure. Temporal data expansions, such as extending time frames, do not alter the CKG's temporal unit structure, avoiding the need for re-embedding entities and relations. Spatial expansions, which might involve modeling multiple adjacent cities, are managed by segmenting study areas according to natural city boundaries defined by geographical and infrastructural characteristics. This potential for further scalability highlights a promising direction for future research.

In real-world scenarios, the integration of the CKG-GNN model into traffic management systems markedly improves our ability to forecast traffic speeds with high accuracy and efficiency. The CKG-GNN model not only surpasses baseline GNN models in predictive accuracy but also maintains a practical inference time of approximately 7 seconds during congested periods, such as the morning peak in $V_{raw}$. The increased accuracy and manageable inference time underscore the model's suitability for real-time applications within traffic management systems, facilitating enhanced operational capabilities at traffic control centers. Specifically, it allows these centers to dynamically adjust traffic signals, optimize routes, and effectively manage traffic flows.
Moreover, the CKG framework deepens our understanding of the dynamic interplay between various contextual factors and traffic systems, providing essential insights for refining traffic management strategies. By including diverse spatial contexts, such as POIs and land uses, the framework empowers traffic management systems to accurately assess traffic conditions under varying land use scenarios. The CKG's temporal dimension illuminates the impacts of traffic congestion and weather conditions on traffic systems, enabling traffic control centers to respond adaptively to sudden traffic jams or severe weather events. This functionality ensures that traffic management systems can leverage real-time, data-driven decisions to significantly enhance traffic regulation and control, optimizing overall traffic flow and safety.

\section{Conclusion}

\subsection{Summary}
This study proposed an innovative context-aware knowledge graph framework designed to organize and embed spatio-temporal contexts through a relation-dependent integration strategy. Integrated with dual-view MHSA and GNN, our CKG-GNN model demonstrates state-of-the-art performance in traffic speed forecasting compared to baseline models.
We find that the optimal KG configuration for the spatial unit was achieved with a buffer of [10-100] meters and [6] spatial links using ComplEx, whereas the temporal unit exhibited the best configuration with a [60] minute buffer and [Hour-Day-Week] temporal links using KG2E. These findings provide a solid foundation for employing a relation-dependent integration strategy to produce context-aware representations and capture the relationships between the traffic system and the surrounding environment.
Incorporating these context-aware representations into traffic forecasting, the CKG-GNN model outperformed benchmarks, achieving an average MAE of 3.46±0.01 and a MAPE of 14.76±0.09\% across speed predictions from 10 to 120 minutes. The feature-view attention heatmap highlighted the dominance of relation-dependent road features over road-only features, asserting the effectiveness of our relation-dependent integration strategy. We also observed the significance of various temporal contexts and the substantial impact of spatial and temporal links on improving forecasting accuracy. Moreover, the sequential-view attention heatmap illustrated the model's capacity to prioritize recent time slots for predicting traffic speed.
These insights affirm the utility of the proposed CKG framework in improving traffic speed forecasting, combining the strengths of knowledge graphs with neural network architectures to advance predictive performance.

\subsection{Limitations and future research}
While this study marks significant strides, it also highlights several limitations that may guide future research directions.
First, optimize KG embedding methods. While our exploration of various KGE techniques was comprehensive, it suggests a potential complexity due to the involvement of both distance-based and similarity-based scoring functions. This complexity could limit their applicability in real-world scenarios. To address this, a future enhancement could be employing AutoML techniques, such as AutoSF \cite{zhang2020autosf}, to autonomously design scoring functions for CKGs and optimize spatio-temporal context configuration. This approach could reduce the complexity associated with evaluating KGE performance and increase the applicability of our framework across diverse scenarios.
Second, explore framework versatility and applicability. While the framework inherently allows for effective integration with various GNN architectures and the inclusion of additional contextual data, its full potential has not been fully explored. Future research endeavors could greatly benefit from conducting a comparative analysis of diverse GNN architectures to thoroughly evaluate and exploit the CKG framework's adaptability and effectiveness across varying models. Furthermore, the integration of additional contextual information during this investigative phase could potentially augment the framework's predictive accuracy and depth of insight for future research.


\bibliographystyle{IEEEtran}
\bibliography{IEEEabrv,ref}

\begin{thebibliography}{10}
\providecommand{\url}[1]{#1}
\csname url@samestyle\endcsname
\providecommand{\newblock}{\relax}
\providecommand{\bibinfo}[2]{#2}
\providecommand{\BIBentrySTDinterwordspacing}{\spaceskip=0pt\relax}
\providecommand{\BIBentryALTinterwordstretchfactor}{4}
\providecommand{\BIBentryALTinterwordspacing}{\spaceskip=\fontdimen2\font plus
\BIBentryALTinterwordstretchfactor\fontdimen3\font minus \fontdimen4\font\relax}
\providecommand{\BIBforeignlanguage}[2]{{%
\expandafter\ifx\csname l@#1\endcsname\relax
\typeout{** WARNING: IEEEtran.bst: No hyphenation pattern has been}%
\typeout{** loaded for the language `#1'. Using the pattern for}%
\typeout{** the default language instead.}%
\else
\language=\csname l@#1\endcsname
\fi
#2}}
\providecommand{\BIBdecl}{\relax}
\BIBdecl

\bibitem{wegener2021land}
M.~Wegener, ``Land-use transport interaction models,'' \emph{Handb. Reg. Sci.}, pp. 229--246, 2021.

\bibitem{zhang2023incorporating}
Y.~Zhang, T.~Zhao, S.~Gao, and M.~Raubal, ``Incorporating multimodal context information into traffic speed forecasting through graph deep learning,'' \emph{Int. J. Geogr. Inf. Sci.}, vol.~37, no.~9, pp. 1909--1935, 2023.

\bibitem{shaygan2022traffic}
M.~Shaygan, C.~Meese, W.~Li, X.~G. Zhao, and M.~Nejad, ``Traffic prediction using artificial intelligence: review of recent advances and emerging opportunities,'' \emph{Transp. Res. Part C Emerg. Technol.}, vol. 145, p. 103921, 2022.

\bibitem{peng2019knowledge}
H.~Peng, N.~Klepp, M.~Toutiaee, I.~B. Arpinar, and J.~A. Miller, ``Knowledge and situation-aware vehicle traffic forecasting,'' in \emph{2019 IEEE Int. Conf. Big Data (Big Data)}.\hskip 1em plus 0.5em minus 0.4em\relax IEEE, 2019, pp. 3803--3812.

\bibitem{yin2021deep}
X.~Yin, G.~Wu, J.~Wei, Y.~Shen, H.~Qi, and B.~Yin, ``Deep learning on traffic prediction: Methods, analysis and future directions,'' \emph{IEEE Trans. Intell. Transp. Syst.}, 2021.

\bibitem{tedjopurnomo2020survey}
D.~A. Tedjopurnomo, Z.~Bao, B.~Zheng, F.~Choudhury, and A.~Qin, ``A survey on modern deep neural network for traffic prediction: Trends, methods and challenges,'' \emph{IEEE Trans. Knowl. Data Eng.}, 2020.

\bibitem{buchin2012context}
M.~Buchin, S.~Dodge, and B.~Speckmann, ``Context-aware similarity of trajectories,'' in \emph{Int. Conf. Geogr. Inf. Sci.}\hskip 1em plus 0.5em minus 0.4em\relax Springer, 2012, pp. 43--56.

\bibitem{zhang2022street}
Y.~Zhang and M.~Raubal, ``Street-level traffic flow and context sensing analysis through semantic integration of multisource geospatial data,'' \emph{Trans. GIS}, vol.~26, no.~8, pp. 3330--3348, 2022.

\bibitem{yu2023survey}
D.~Yu, B.~Yang, D.~Liu, H.~Wang, and S.~Pan, ``A survey on neural-symbolic learning systems,'' \emph{Neural Networks}, 2023.

\bibitem{kang2012intra}
C.~Kang, X.~Ma, D.~Tong, and Y.~Liu, ``Intra-urban human mobility patterns: An urban morphology perspective,'' \emph{Phys. A Stat. Mech. its Appl.}, vol. 391, no.~4, pp. 1702--1717, 2012.

\bibitem{wang2021spatio}
H.~Wang, Q.~Yu, Y.~Liu, D.~Jin, and Y.~Li, ``Spatio-temporal urban knowledge graph enabled mobility prediction,'' \emph{Proc. ACM Interactive, Mobile, Wearable Ubiquitous Technol.}, vol.~5, no.~4, pp. 1--24, 2021.

\bibitem{zhang2023mobility}
Q.~Zhang, Z.~Ma, P.~Zhang, and E.~Jenelius, ``Mobility knowledge graph: review and its application in public transport,'' \emph{Transportation}, pp. 1--27, 2023.

\bibitem{ahmed2022knowledge}
U.~Ahmed, G.~Srivastava, Y.~Djenouri, and J.~C.-W. Lin, ``Knowledge graph based trajectory outlier detection in sustainable smart cities,'' \emph{Sustain. Cities Soc.}, vol.~78, p. 103580, 2022.

\bibitem{dai2020survey}
Y.~Dai, S.~Wang, N.~N. Xiong, and W.~Guo, ``A survey on knowledge graph embedding: Approaches, applications and benchmarks,'' \emph{Electronics}, vol.~9, no.~5, p. 750, 2020.

\bibitem{chen2020review}
X.~Chen, S.~Jia, and Y.~Xiang, ``A review: Knowledge reasoning over knowledge graph,'' \emph{Expert Syst. Appl.}, vol. 141, p. 112948, 2020.

\bibitem{cao2024knowledge}
J.~Cao, J.~Fang, Z.~Meng, and S.~Liang, ``Knowledge graph embedding: A survey from the perspective of representation spaces,'' \emph{ACM Computing Surveys}, vol.~56, no.~6, pp. 1--42, 2024.

\bibitem{wang2017knowledge}
Q.~Wang, Z.~Mao, B.~Wang, and L.~Guo, ``Knowledge graph embedding: A survey of approaches and applications,'' \emph{IEEE Trans. Knowl. Data Eng.}, vol.~29, no.~12, pp. 2724--2743, 2017.

\bibitem{zhang2024knowledge}
X.~Han, X.~Zhang, Y.~Wu, Z.~Zhang, T.~Zhang, and Y.~Wang, ``Knowledge-based multiple relations modeling for traffic forecasting,'' \emph{IEEE Transactions on Intelligent Transportation Systems}, vol.~25, no.~9, pp. 11\,844--11\,857, 2024.

\bibitem{chadzynski2022semantic}
A.~Chadzynski, S.~Li, A.~Grisiute, F.~Farazi, C.~Lindberg, S.~Mosbach, P.~Herthogs, and M.~Kraft, ``Semantic 3d city agents—an intelligent automation for dynamic geospatial knowledge graphs,'' \emph{Energy AI}, vol.~8, p. 100137, 2022.

\bibitem{zhuang2017understanding}
C.~Zhuang, N.~J. Yuan, R.~Song, X.~Xie, and Q.~Ma, ``Understanding people lifestyles: Construction of urban movement knowledge graph from gps trajectory.'' in \emph{IJCAI}, 2017, pp. 3616--3623.

\bibitem{zhu2022kst}
J.~Zhu, X.~Han, H.~Deng, C.~Tao, L.~Zhao, P.~Wang, T.~Lin, and H.~Li, ``Kst-gcn: A knowledge-driven spatial-temporal graph convolutional network for traffic forecasting,'' \emph{IEEE Trans. Intell. Transp. Syst.}, 2022.

\bibitem{rossi2021knowledge}
A.~Rossi, D.~Barbosa, D.~Firmani, A.~Matinata, and P.~Merialdo, ``Knowledge graph embedding for link prediction: A comparative analysis,'' \emph{ACM Trans. Knowl. Discov. from Data}, vol.~15, no.~2, pp. 1--49, 2021.

\bibitem{he2015learning}
S.~He, K.~Liu, G.~Ji, and J.~Zhao, ``Learning to represent knowledge graphs with gaussian embedding,'' in \emph{Proc. 24th ACM Int. Conf. Inf. Knowl. Manag.}, 2015, pp. 623--632.

\bibitem{kumar2021applications}
N.~Kumar and M.~Raubal, ``Applications of deep learning in congestion detection, prediction and alleviation: A survey,'' \emph{Transp. Res. Part C Emerg. Technol.}, vol. 133, p. 103432, 2021.

\bibitem{feng2023macro}
S.~Feng, S.~Wei, J.~Zhang, Y.~Li, J.~Ke, G.~Chen, Y.~Zheng, and H.~Yang, ``A macro--micro spatio-temporal neural network for traffic prediction,'' \emph{Transp. Res. Part C Emerg. Technol.}, vol. 156, p. 104331, 2023.

\bibitem{li2017diffusion}
Y.~Li, R.~Yu, C.~Shahabi, and Y.~Liu, ``Diffusion convolutional recurrent neural network: Data-driven traffic forecasting,'' \emph{arXiv preprint arXiv:1707.01926}, 2017.

\bibitem{yu2017spatio}
B.~Yu, H.~Yin, and Z.~Zhu, ``Spatio-temporal graph convolutional networks: A deep learning framework for traffic forecasting,'' \emph{arXiv preprint arXiv:1709.04875}, 2017.

\bibitem{zhao2019t}
L.~Zhao, Y.~Song, C.~Zhang, Y.~Liu, P.~Wang, T.~Lin, M.~Deng, and H.~Li, ``T-gcn: A temporal graph convolutional network for traffic prediction,'' \emph{IEEE Trans. Intell. Transp. Syst.}, vol.~21, no.~9, pp. 3848--3858, 2019.

\bibitem{zhu2021ast}
J.~Zhu, Q.~Wang, C.~Tao, H.~Deng, L.~Zhao, and H.~Li, ``Ast-gcn: Attribute-augmented spatiotemporal graph convolutional network for traffic forecasting,'' \emph{IEEE Access}, vol.~9, pp. 35\,973--35\,983, 2021.

\bibitem{ali2021pykeen}
M.~Ali, M.~Berrendorf, C.~T. Hoyt, L.~Vermue, S.~Sharifzadeh, V.~Tresp, and J.~Lehmann, ``Pykeen 1.0: a python library for training and evaluating knowledge graph embeddings,'' \emph{J. Mach. Learn. Res.}, vol.~22, no.~82, pp. 1--6, 2021.

\bibitem{hoyt2022unified}
C.~T. Hoyt, M.~Berrendorf, M.~Galkin, V.~Tresp, and B.~M. Gyori, ``A unified framework for rank-based evaluation metrics for link prediction in knowledge graphs,'' \emph{arXiv preprint arXiv:2203.07544}, 2022.

\bibitem{wang2021libcity}
J.~Wang, J.~Jiang, W.~Jiang, C.~Li, and W.~X. Zhao, ``Libcity: An open library for traffic prediction,'' in \emph{Proc. 29th Int. Conf. Adv. Geogr. Inf. Syst.}, 2021, pp. 145--148.

\bibitem{gneiting2007strictly}
T.~Gneiting and A.~E. Raftery, ``Strictly proper scoring rules, prediction, and estimation,'' \emph{Journal of the American statistical Association}, vol. 102, no. 477, pp. 359--378, 2007.

\bibitem{lana2024measuring}
I.~Laña, I.~Olabarrieta, and J.~D. Ser, ``Measuring the confidence of single-point traffic forecasting models: Techniques, experimental comparison, and guidelines toward their actionability,'' \emph{IEEE Transactions on Intelligent Transportation Systems}, vol.~25, no.~9, pp. 11\,180--11\,199, 2024.

\bibitem{zhang2020autosf}
Y.~Zhang, Q.~Yao, W.~Dai, and L.~Chen, ``Autosf: Searching scoring functions for knowledge graph embedding,'' in \emph{2020 IEEE 36th Int. Conf. Data Eng.}\hskip 1em plus 0.5em minus 0.4em\relax IEEE, 2020, pp. 433--444.

\end{thebibliography}

\vspace{-33pt}
\begin{IEEEbiographynophoto}{Yatao Zhang}
received the B.S. degree and the M.S. degrees in geographical information science from Sun Yat-sen University, Guangzhou, China, in 2017 and Wuhan University, Wuhan, China, in 2020, respectively. He is a doctoral student at the Mobility Information Engineering lab at ETH Zurich and the Future Resilient Systems at the Singapore-ETH Center. His research interests lie in context-based spatiotemporal analysis, geospatial big data mining, and traffic forecasting.
\end{IEEEbiographynophoto}

\vspace{-33pt}
\begin{IEEEbiographynophoto}{Yi Wang}
received the Ph.D. degree in transportation engineering from the University of Hong Kong, Hong Kong, China. She is a lecturer at the Civil and Natural Resources Engineering Department, University of Canterbury. Her research interests lie in transport planning with a focus on sustainability and resilience, which involves developing modeling frameworks and solution methods.
\end{IEEEbiographynophoto}

\vspace{-33pt}
\begin{IEEEbiographynophoto}{Song Gao}
received the Ph.D. degree in GIScience from the University of California, Santa Barbara, USA, in 2017. He is an associate professor in GIScience at the Department of Geography, University of Wisconsin-Madison. His main research interests include place-based GIS, geospatial data science, and GeoAI approaches to human mobility and social sensing.
\end{IEEEbiographynophoto}

\vspace{-33pt}
\begin{IEEEbiographynophoto}{Martin Raubal}
received the Ph.D. degree in Geoinformation from Vienna University of Technology, Vienna, Austria, in 2001. He is a professor of geoinformation engineering at ETH Zurich. His research interests focus on spatial decision-making for sustainability, more specifically, he concentrates on analyzing spatio-temporal aspects of human mobility, spatial cognitive engineering, and mobile eye-tracking to investigate visual attention while interacting with geoinformation and in spatial decision situations.
\end{IEEEbiographynophoto}


\clearpage
\begin{appendices}
\setcounter{table}{0}
\renewcommand{\thetable}{A\arabic{table}}
\renewcommand{\theHtable}{A\arabic{table}}

\section*{Appendix}

\subsection{Comparison of KGE techniques}

Table \ref{tab:kge_tech} provides an overview of the scoring functions and key features of various KGE techniques, encompassing both distance-based and similarity-based models.


\begin{table*}[hbp!]
\centering
\textcolor{mycolor}{
\caption{Comparison of key features and differences among various knowledge graph embedding techniques.}
\label{tab:kge_tech}
\renewcommand\arraystretch{1.0}
\makebox[\textwidth]{
\begin{tabular}{p{1cm}llp{8cm}}
\hline
\multicolumn{1}{l}{Category}      & Method  & Scoring functions                                                                                                                                                                                                                                                                        & Key Features                                                                                                                  \\ \hline
\multirow{3}{*}{\shortstack[l]{Distance- \\ based}}   & TransE  & $-\left \| \textbf{h} +   \textbf{r} - \textbf{t} \right \|^2_2$                                                                                                                                                                                                                        & Represents entities and relations in the same vector space; struggles with complex relation types.                            \\
                                  & TransR  & $-\left \| \hat{\textbf{h}} +   \textbf{r} - \hat{\textbf{t}} \right \|^2_2$                                                                                                                                                                                                            & Introduces relation-specific embedding spaces; enhances handling of diverse relation types compared to TransE.                \\
                                  & KG2E    & $-\text{tr}(\boldsymbol{\Sigma}_r^{-1}   (\boldsymbol{\Sigma}_h + \boldsymbol{\Sigma}_t)) - \boldsymbol{\mu}^T   \boldsymbol{\Sigma}_r^{-1} \boldsymbol{\mu} -   \ln\left(\frac{\text{det}(\boldsymbol{\Sigma}_r)}{\text{det}(\boldsymbol{\Sigma}_h   + \boldsymbol{\Sigma}_t)}\right)$ & Models entities and relations with multi-variate Gaussian distributions; captures inherent uncertainties in knowledge graphs. \\ \hline
\multirow{3}{*}{\shortstack[l]{Similarity- \\based}} & RESCAL  & $\sum_{i=0}^{e-1}\sum_{j=0}^{e-1}[\textbf{M}_r]_{ij}   \cdot [\textbf{h}]_i \cdot [\textbf{t}]_j$                                                                                                                                                                                       & Associates each entity with a vector and each relation with a matrix; leads to a surge in the number of parameters.           \\
                                  & ComplEx & $Real\left ( \sum_{i=0}^{e-1}   [\textbf{r}]_i \cdot [\textbf{h}]_i \cdot [\bar{\textbf{t}}]_i \right )$                                                                                                                                                                                & Utilizes complex-valued embeddings for entities and relations; effectively models asymmetric relationships.                   \\
                                  & NTN     & $\textbf{\textbf{r}}^T   \text{tanh}\left ( \textbf{h}^T \dot{\textbf{M}}_r \textbf{t} +   \textbf{M}_r^1 \textbf{h} + \textbf{M}_r^2 \textbf{t} + \textbf{b}_r\right )$                                                                                                                & Employes a tensor-based architecture; improves relational modeling but requires a substantial number of parameters.           \\ \hline
\end{tabular}
}
}
\end{table*}

\subsection{Traffic speed and context datasets}


\begin{table}[htbp!]
\centering
\textcolor{mycolor}{
\caption{Basic information for traffic speed and context datasets.}
\label{tab:dataset}
\renewcommand\arraystretch{1.0}
\begin{tabular}{p{0.8cm}p{1.5cm}p{2.3cm}p{2.3cm}}
\hline
Datasets                                               & Time Frame                                                                           & Resolutions                                                                             & Data sources                                                                               \\ \hline
$V_{raw}$                                              & \begin{tabular}[c]{@{}l@{}}06:00-12:00, \\ 08-25/03/2022\end{tabular}                & \begin{tabular}[c]{@{}l@{}}10 min per road \\ segment (resample)\end{tabular}          & HERE API                                                                             \\
$V_{work}$                                             & \begin{tabular}[c]{@{}l@{}}00:00-23:59, \\ 23/03/2022\end{tabular}                   & \begin{tabular}[c]{@{}l@{}}10 min per road \\ segment (resample)\end{tabular}          & HERE API                                                                             \\
$V_{rest}$                                             & \begin{tabular}[c]{@{}l@{}}00:00-23:59, \\ 19/03/2022\end{tabular}                   & \begin{tabular}[c]{@{}l@{}}10 min per road \\ segment (resample)\end{tabular}          & HERE API                                                                             \\ \hline
Road                                                   & 2022                                                                                 & \begin{tabular}[c]{@{}l@{}}1,606 road segments\\  (Sinagpore CCR)\end{tabular}         & HERE API                                                                             \\
POI                                                    & 2022                                                                                 & \begin{tabular}[c]{@{}l@{}}13 categories, 85,647\\  points (Sinagpore)\end{tabular}    & \begin{tabular}[c]{@{}l@{}}OneMap, DataMall, \\ OpenStreetMap\end{tabular}           \\
\begin{tabular}[c]{@{}l@{}}Land \\ use\end{tabular}    & 2019                                                                                 & \begin{tabular}[c]{@{}l@{}}28 categories, 113,212\\  polygons (Singapore)\end{tabular} & \begin{tabular}[c]{@{}l@{}}Urban Redevelopment \\ Authority (Singapore)\end{tabular} \\
\begin{tabular}[c]{@{}l@{}}Jam \\ factor\end{tabular}  & \begin{tabular}[c]{@{}l@{}}Same as $V_{raw}$, \\ $V_{work}$, $V_{rest}$\end{tabular} & \begin{tabular}[c]{@{}l@{}}10 min per road \\ segment (resample)\end{tabular}          & HERE API                                                                             \\
\begin{tabular}[c]{@{}l@{}}Weather\\ data\end{tabular} & \begin{tabular}[c]{@{}l@{}}Same as $V_{raw}$, \\ $V_{work}$, $V_{rest}$\end{tabular} & \begin{tabular}[c]{@{}l@{}}10 min per weather \\ station (resample)\end{tabular}       & \begin{tabular}[c]{@{}l@{}}National Environment \\ Agency (Singapore)\end{tabular}   \\ \hline
\end{tabular}
}
\end{table}

\subsection{Additional KGE evaluation metrics}
Table \ref{tab:spat_embed_buffer_apd} supplements Table \ref{tab:spat_embed_buffer} with additional both-side KGE evaluation metrics for embedding $\mathbb{K}_S$ under two spatial link configurations: [-] indicating no spatial link and [6] indicating Link[6]. These metrics, including MRR, H@5, and AH@K, also demonstrate that the best performance of embedding $\mathbb{K}_S$ is achieved using ComplEx with Buffer[10-100] and Link[6].

\begin{table}[htbp!]
\centering
\textcolor{mycolor}{
\caption{Additional both-side KGE evaluation metrics of $\mathbb{K}_S$ with different configurations of buffer [Dist].}
\label{tab:spat_embed_buffer_apd}
\renewcommand\arraystretch{1.0}
\begin{tabular}{p{0.8cm}p{0.8cm}p{0.6cm}p{0.6cm}p{0.6cm}p{0.6cm}p{0.6cm}p{0.6cm}} 
\hline
\multirow{2}{*}{Model}   & \multirow{2}{*}{Metrics} & \multicolumn{2}{c}{Buffer{[}10-100{]}} & \multicolumn{2}{c}{Buffer{[}100-500{]}} & \multicolumn{2}{c}{Buffer{[}10-500{]}} \\ \cline{3-8} 
                         &                          & {[}-{]}        & {[}6{]}       & {[}-{]}        & {[}6{]}        & {[}-{]}        & {[}6{]}       \\ \hline
\multirow{3}{*}{TransE}  & MRR                      & \textbf{0.43}      & 0.57              & \textbf{0.21}      & 0.49               & 0.27               & 0.45              \\
                         & H@5                      & \textbf{0.55}      & 0.71              & \textbf{0.30}      & 0.60               & 0.36               & 0.57              \\
                         & AH@K                     & \textbf{0.63}      & 0.80              & \textbf{0.41}      & 0.70               & 0.45               & 0.67              \\ \cline{2-8} 
\multirow{3}{*}{TransR}  & MRR                      & 0.34               & 0.56              & \textbf{0.21}      & 0.46               & 0.25               & 0.44              \\
                         & H@5                      & 0.47               & 0.69              & 0.29               & 0.58               & 0.35               & 0.56              \\
                         & AH@K                     & 0.56               & 0.79              & 0.39               & 0.68               & 0.45               & 0.66              \\ \cline{2-8} 
\multirow{3}{*}{KG2E}    & MRR                      & 0.40               & 0.50              & \textbf{0.21}      & 0.42               & 0.26               & 0.43              \\
                         & H@5                      & 0.52               & 0.63              & 0.29               & 0.53               & 0.35               & 0.55              \\
                         & AH@K                     & 0.60               & 0.73              & 0.39               & 0.63               & 0.44               & 0.64              \\ \hline
\multirow{3}{*}{RESCAL}  & MRR                      & 0.01               & 0.00              & 0.01               & 0.00               & 0.01               & 0.00              \\
                         & H@5                      & 0.00               & 0.00              & 0.00               & 0.00               & 0.00               & 0.00              \\
                         & AH@K                     & 0.00               & 0.00              & 0.00               & 0.00               & 0.00               & 0.00              \\ \cline{2-8} 
\multirow{3}{*}{ComplEx} & MRR                      & 0.40               & \textbf{0.68}     & 0.20               & \textbf{0.55}      & \textbf{0.28}      & \textbf{0.53}     \\
                         & H@5                      & 0.53               & \textbf{0.82}     & 0.28               & \textbf{0.68}      & \textbf{0.37}      & \textbf{0.66}     \\
                         & AH@K                     & 0.62               & \textbf{0.89}     & 0.40               & \textbf{0.76}      & \textbf{0.47}      & \textbf{0.74}     \\ \cline{2-8} 
\multirow{3}{*}{NTN}     & MRR                      & 0.05               & 0.29              & 0.09               & 0.17               & 0.11               & 0.15              \\
                         & H@5                      & 0.07               & 0.51              & 0.14               & 0.30               & 0.18               & 0.26              \\
                         & AH@K                     & 0.11               & 0.66              & 0.21               & 0.50               & 0.24               & 0.46              \\ \hline
\end{tabular}
}
\end{table}

Table \ref{tab:spat_embed_link_apd} extends Table \ref{tab:spat_embed_link} by providing additional both-side KGE evaluation metrics for embedding $\mathbb{K}_S$ with various spatial links. [-] indicates no spatial link, while [1,...,6] corresponds to spatial links with [Order] ranging from 1 to 6. The metrics MRR, H@5, and AH@K further confirm that ComplEx consistently outperforms other methods across most configurations. Also, the embedding performance improves with the inclusion of spatial links.

\begin{table}[htbp!]
\centering
\textcolor{mycolor}{
\caption{Additional both-side KGE evaluation metrics of embedding $\mathbb{K}_S$ across various spatial links.}
\label{tab:spat_embed_link_apd}
\renewcommand\arraystretch{1.0}
\begin{tabular}{p{0.8cm}p{0.6cm}p{0.6cm}p{0.4cm}p{0.4cm}p{0.4cm}p{0.4cm}p{0.4cm}p{0.4cm}p{0.4cm}} 
\hline
Model                    & Buffer                         & Metrics & {[}-{]}   & {[}1{]}   & {[}2{]}   & {[}3{]}   & {[}4{]}   & {[}5{]}   & {[}6{]}   \\ \hline
\multirow{3}{*}{ComplEx} & \multirow{6}{*}{\parbox{6cm}{[10-\\100]}}  & MRR     & 0.40          & \textbf{0.42} & \textbf{0.44} & \textbf{0.48} & \textbf{0.55} & \textbf{0.63} & \textbf{0.68} \\
                         &                                & H@5     & 0.53          & \textbf{0.56} & \textbf{0.60} & \textbf{0.67} & \textbf{0.73} & \textbf{0.79} & \textbf{0.82} \\
                         &                                & AH@K    & 0.62          & \textbf{0.64} & \textbf{0.68} & \textbf{0.75} & \textbf{0.81} & \textbf{0.86} & \textbf{0.89} \\ \cline{3-10} 
\multirow{3}{*}{TransE}  &                                & MRR     & \textbf{0.43} & \textbf{0.42} & 0.42          & 0.45          & 0.49          & 0.54          & 0.57          \\
                         &                                & H@5     & \textbf{0.55} & \textbf{0.56} & 0.58          & 0.62          & 0.66          & 0.70          & 0.71          \\
                         &                                & AH@K    & \textbf{0.63} & \textbf{0.64} & 0.67          & 0.71          & 0.76          & 0.79          & 0.80          \\ \hline
\multirow{3}{*}{ComplEx} & \multirow{6}{*}{\parbox{6cm}{[100-\\500]}} & MRR     & 0.20          & \textbf{0.22} & \textbf{0.24} & \textbf{0.29} & \textbf{0.38} & \textbf{0.47} & \textbf{0.55} \\
                         &                                & H@5     & 0.28          & \textbf{0.31} & \textbf{0.34} & \textbf{0.41} & \textbf{0.51} & \textbf{0.60} & \textbf{0.68} \\
                         &                                & AH@K    & 0.40          & \textbf{0.43} & \textbf{0.46} & \textbf{0.53} & \textbf{0.61} & \textbf{0.70} & \textbf{0.76} \\ \cline{3-10} 
\multirow{3}{*}{TransE}  &                                & MRR     & \textbf{0.21} & \textbf{0.22} & \textbf{0.24} & 0.28          & 0.35          & 0.42          & 0.49          \\
                         &                                & H@5     & \textbf{0.30} & \textbf{0.31} & \textbf{0.34} & 0.39          & 0.47          & 0.55          & 0.60          \\
                         &                                & AH@K    & \textbf{0.41} & 0.42          & 0.45          & 0.50          & 0.58          & 0.65          & 0.70          \\ \hline
\multirow{3}{*}{ComplEx} & \multirow{6}{*}{\parbox{6cm}{[10-\\500]}}  & MRR     & \textbf{0.28} & \textbf{0.28} & \textbf{0.29} & \textbf{0.33} & \textbf{0.39} & \textbf{0.46} & \textbf{0.53} \\
                         &                                & H@5     & \textbf{0.37} & \textbf{0.37} & \textbf{0.39} & \textbf{0.45} & \textbf{0.52} & \textbf{0.59} & \textbf{0.66} \\
                         &                                & AH@K    & \textbf{0.47} & \textbf{0.48} & \textbf{0.50} & \textbf{0.55} & \textbf{0.61} & \textbf{0.68} & \textbf{0.74} \\ \cline{3-10} 
\multirow{3}{*}{TransE}  &                                & MRR     & 0.27          & 0.26          & 0.28          & 0.29          & 0.34          & 0.39          & 0.45          \\
                         &                                & H@5     & 0.36          & 0.36          & 0.38          & 0.41          & 0.46          & 0.52          & 0.57          \\
                         &                                & AH@K    & 0.45          & 0.45          & 0.48          & 0.51          & 0.56          & 0.62          & 0.67          \\ \hline
\end{tabular}
}
\end{table}

In addition to Table \ref{tab:temp_embed_time}, Table \ref{tab:temp_embed_time_apd} presents additional evaluation metrics (MRR, H@5, and AH@K) for different [PastMins] configurations under two temporal link scenarios: [-] indicating no temporal links and [HDW] incorporating hourly, daily, and weekly links. These metrics reveal that while MRR suggests ComplEx performs well, both H@5 and AH@K still highlight KG2E's superior performance. This distinction stems from the metrics focusing on different aspects of model effectiveness: MRR emphasizes immediate accuracy—crucial in recommendation systems, whereas H@5 and AH@K prioritize overall quality and consistency—essential for our framework. Thus, we have chosen KG2E for our knowledge graph embedding to ensure reliable and robust performance.

In addition to Table \ref{tab:temp_embed_link}, Table \ref{tab:temp_embed_link_apd} presents additional evaluation metrics (MRR, H@5, and AH@K) using KG2E to embed $\mathbb{K}_T$ across various [PastMins] configurations. These results notably underscore the superior performance of the Link[HDW] and Past[60] configuration.

\begin{table*}[htbp!]
\centering
\textcolor{mycolor}{
\caption{Additional both-side KGE evaluation metrics of embedding $\mathbb{K}_T$ with different configurations of [PastMins].}
\label{tab:temp_embed_time_apd}
\renewcommand\arraystretch{1.0}
\makebox[\textwidth]{
\begin{tabular}{cccccccccccccc}
\hline
\multirow{2}{*}{Model}   & \multirow{2}{*}{Metrics} & \multicolumn{2}{c}{Past{[}10{]}} & \multicolumn{2}{c}{Past{[}20{]}} & \multicolumn{2}{c}{Past{[}30{]}} & \multicolumn{2}{c}{Past{[}40{]}} & \multicolumn{2}{c}{Past{[}50{]}} & \multicolumn{2}{c}{Past{[}60{]}} \\ \cline{3-14} 
                         &                          & {[}-{]}         & {[}HDW{]}      & {[}-{]}         & {[}HDW{]}      & {[}-{]}         & {[}HDW{]}      & {[}-{]}         & {[}HDW{]}      & {[}-{]}         & {[}HDW{]}      & {[}-{]}         & {[}HDW{]}      \\ \hline
\multirow{3}{*}{TransE}  & MRR                      & 0.29            & 0.45           & 0.33            & 0.45           & 0.37            & 0.47           & 0.45            & 0.47           & 0.45            & 0.53           & 0.47            & 0.53           \\
                         & H@5                      & 0.48            & 0.70           & 0.56            & 0.70           & 0.60            & 0.73           & 0.70            & 0.70           & 0.70            & 0.77           & 0.73            & 0.77           \\
                         & AH@K                     & 0.89            & 0.97           & 0.90            & 0.97           & 0.93            & 0.98           & 0.97            & 0.98           & 0.97            & 1.00           & 0.98            & 0.99           \\ \cline{2-14} 
\multirow{3}{*}{TransR}  & MRR                      & \textbf{0.55}   & 0.49           & \textbf{0.50}   & 0.47           & 0.50            & 0.47           & 0.49            & 0.46           & 0.47            & 0.46           & 0.47            & 0.46           \\
                         & H@5                      & \textbf{0.56}   & 0.57           & 0.48            & 0.62           & 0.59            & 0.69           & 0.57            & 0.67           & 0.62            & 0.71           & 0.69            & 0.74           \\
                         & AH@K                     & \textbf{0.94}   & 0.89           & 0.88            & 0.85           & 0.91            & 0.86           & 0.89            & 0.84           & 0.85            & 0.85           & 0.86            & 0.86           \\ \cline{2-14} 
\multirow{3}{*}{KG2E}    & MRR                      & 0.38            & 0.55           & 0.49            & 0.56           & \textbf{0.54}   & 0.56           & 0.55            & 0.60           & 0.56            & 0.62           & 0.56            & 0.66           \\
                         & H@5                      & 0.49            & \textbf{0.84}  & \textbf{0.66}   & \textbf{0.87}  & \textbf{0.77}   & 0.82           & \textbf{0.84}   & \textbf{0.90}  & \textbf{0.87}   & \textbf{0.94}  & \textbf{0.82}   & \textbf{0.95}  \\
                         & AH@K                     & 0.89            & \textbf{0.98}  & \textbf{0.97}   & \textbf{0.99}  & \textbf{0.97}   & \textbf{0.99}  & \textbf{0.98}   & \textbf{0.99}  & \textbf{0.99}   & \textbf{0.99}  & \textbf{0.99}   & \textbf{1.00}  \\ \hline
\multirow{3}{*}{RESCAL}  & MRR                      & 0.28            & 0.24           & 0.25            & 0.23           & 0.24            & 0.23           & 0.24            & 0.24           & 0.23            & 0.22           & 0.23            & 0.22           \\
                         & H@5                      & 0.25            & 0.27           & 0.30            & 0.27           & 0.27            & 0.28           & 0.27            & 0.27           & 0.27            & 0.27           & 0.28            & 0.28           \\
                         & AH@K                     & -0.01           & -0.01          & -0.01           & -0.01          & -0.01           & -0.01          & -0.01           & -0.01          & -0.01           & -0.01          & -0.01           & -0.01          \\ \cline{2-14} 
\multirow{3}{*}{ComplEx} & MRR                      & 0.27            & \textbf{0.63}  & 0.47            & \textbf{0.66}  & 0.40            & \textbf{0.70}  & \textbf{0.61}   & \textbf{0.72}  & \textbf{0.67}   & \textbf{0.73}  & \textbf{0.71}   & \textbf{0.72}  \\
                         & H@5                      & 0.40            & 0.72           & 0.50            & 0.78           & 0.58            & \textbf{0.83}  & 0.67            & 0.86           & 0.78            & 0.86           & \textbf{0.82}   & 0.86           \\
                         & AH@K                     & -0.01           & 0.57           & 0.01            & 0.72           & 0.24            & 0.79           & 0.48            & 0.84           & 0.69            & 0.85           & 0.77            & 0.85           \\ \cline{2-14} 
\multirow{3}{*}{NTN}     & MRR                      & 0.15            & 0.17           & 0.17            & 0.16           & 0.16            & 0.16           & 0.17            & 0.17           & 0.16            & 0.16           & 0.16            & 0.16           \\
                         & H@5                      & 0.30            & 0.30           & 0.29            & 0.28           & 0.30            & 0.29           & 0.30            & 0.29           & 0.28            & 0.29           & 0.29            & 0.30           \\
                         & AH@K                     & 0.00            & 0.00           & 0.00            & 0.01           & 0.01            & 0.01           & 0.00            & 0.02           & 0.01            & 0.04           & 0.01            & 0.05           \\ \hline
\end{tabular}
}
}
\end{table*}

\begin{table}[htbp!]
\centering
\textcolor{mycolor}{
\caption{Additional both-side KGE evaluation metrics with various temporal links when embedding $\mathbb{K}_T$.}
\label{tab:temp_embed_link_apd}
\renewcommand\arraystretch{1.0}
\begin{tabular}{p{0.5cm}p{0.5cm}p{0.6cm}p{0.6cm}p{0.6cm}p{0.6cm}p{0.6cm}p{0.6cm}p{0.7cm}} 
\hline
Model                 & Link                        & Metrics & Past{[}10{]}  & Past{[}20{]}  & Past{[}30{]}  & Past{[}40{]}  & Past{[}50{]}  & Past{[}60{]}  \\ \hline
\multirow{3}{*}{KG2E} & \multirow{3}{*}{{[}-{]}}    & MRR     & 0.38          & 0.49          & 0.54          & 0.55          & 0.56          & 0.56          \\
                      &                             & H@5     & 0.49          & 0.66          & 0.77          & 0.84          & 0.87          & 0.82          \\
                      &                             & AH@K    & 0.89          & 0.97          & 0.97          & 0.98          & \textbf{0.99} & 0.99          \\ \hline
\multirow{3}{*}{KG2E} & \multirow{3}{*}{{[}hour{]}} & MRR     & 0.49          & 0.54          & 0.55          & 0.56          & 0.56          & 0.60          \\
                      &                             & H@5     & 0.66          & 0.77          & \textbf{0.84} & 0.87          & 0.82          & 0.90          \\
                      &                             & AH@K    & 0.97          & 0.97          & 0.98          & \textbf{0.99} & \textbf{0.99} & 0.99          \\ \hline
\multirow{3}{*}{KG2E} & \multirow{3}{*}{{[}day{]}}  & MRR     & 0.49          & 0.54          & 0.55          & 0.56          & 0.56          & 0.60          \\
                      &                             & H@5     & 0.66          & 0.77          & \textbf{0.84} & 0.87          & 0.82          & 0.90          \\
                      &                             & AH@K    & 0.97          & 0.97          & 0.98          & \textbf{0.99} & \textbf{0.99} & 0.99          \\ \hline
\multirow{3}{*}{KG2E} & \multirow{3}{*}{{[}week{]}} & MRR     & 0.49          & 0.54          & 0.55          & 0.56          & 0.56          & 0.60          \\
                      &                             & H@5     & 0.66          & 0.77          & \textbf{0.84} & 0.87          & 0.82          & 0.90          \\
                      &                             & AH@K    & 0.97          & 0.97          & 0.98          & \textbf{0.99} & \textbf{0.99} & 0.99          \\ \hline
\multirow{3}{*}{KG2E} & \multirow{3}{*}{{[}HDW{]}}  & MRR     & \textbf{0.55} & \textbf{0.56} & \textbf{0.56} & \textbf{0.6}  & \textbf{0.62} & \textbf{0.66} \\
                      &                             & H@5     & \textbf{0.84} & \textbf{0.87} & 0.82          & \textbf{0.9}  & \textbf{0.94} & \textbf{0.95} \\
                      &                             & AH@K    & \textbf{0.98} & \textbf{0.99} & \textbf{0.99} & \textbf{0.99} & \textbf{0.99} & \textbf{1.00} \\ \hline
\end{tabular}
}
\end{table}

\subsection{Mean scaled interval score}\label{app:msis}
The mean scaled interval score (MSIS), derived from the interval score, evaluates the uncertainty level of forecasting intervals by measuring both the width of the prediction intervals and penalizing when the observed value lies outside the predicted range \cite{lana2024measuring, gneiting2007strictly}. The MSIS equation is defined as follows:
\begin{equation}
    \begin{split}
        \text{MSIS} = & \frac{1}{h} \sum_{t=1}^{h} \left( (U_t - L_t) + \frac{2}{\alpha} (L_t - Y_t) \mathbb{1}(Y_t < L_t) \right. \\
        & \left. + \frac{2}{\alpha} (Y_t - U_t) \mathbb{1}(Y_t > U_t) \right) \cdot \frac{1}{\frac{1}{n} \sum_{t=1}^{n} Y_t}
    \end{split}
\end{equation}

Here, $h$ represents the forecast time horizon. $U_t$ and $L_t$ denote the upper and lower bounds of the prediction interval at time step $t$, respectively, while $Y_t$ refers to the observed value at time $t$. These bounds were obtained using the bootstrap method with 1000 resamples of our model runs. The parameter $\alpha$ represents the quantile, set at 0.05 to provide a 95\% prediction interval. The indicator function, $\mathbb{1}(\cdot)$, takes the value of 1 when the observed value lies outside the prediction interval and 0 otherwise. Finally, $\frac{1}{\frac{1}{n} \sum_{t=1}^{n} Y_t}$ serves as the scale, which is calculated based on the mean of the truth values and is used to normalize the score relative to the overall magnitude of the observed data. Different from the mean absolute difference between time steps, this scale provides a constant reference for normalization that reflects the overall level of the dataset.

\end{appendices}

\end{document}